\pdfoutput=1

\documentclass[11pt]{article}
\usepackage{colortbl}
\usepackage{mdwlist}
\usepackage{booktabs}
\usepackage{makecell}
\usepackage{acl}

\newif\ifcomment\commenttrue
\usepackage[a-1b]{pdfx}

\usepackage{tikz}
\usetikzlibrary{shapes,decorations, fpu}

\usepackage{framed}
\usepackage{mdwlist}
\usepackage{siunitx}
\usepackage{latexsym}
\usepackage{colortbl}
\usepackage{soul, color}
\usepackage{xcolor}
\usepackage{nicefrac}
\usepackage{booktabs}
\usepackage{fnpct}
\usepackage{amsfonts}
\usepackage[T1]{fontenc}
\usepackage{bold-extra}
\usepackage{amsmath}
\usepackage{amssymb}
\usepackage{bm}
\usepackage{graphicx}
\usepackage{mathtools}
\usepackage{microtype}
\usepackage{multirow}
\usepackage{multicol}
\usepackage{xpatch}
\usepackage{latexsym,comment}
\usepackage[normalem]{ulem}

\newcommand*{\missingreference}{{\Huge \colorbox{red}{?reference?}}}
\newcommand*{\missingcitation}{{\Huge \colorbox{red}{?citation?}}}

\makeatletter
\xpatchcmd{\@setref}{\bfseries}{\missingreference}{}{}
\def\@citex[#1]#2{\leavevmode
    \let\@citea\@empty
    \@cite{\@for\@citeb:=#2\do
        {\@citea\def\@citea{,\penalty\@m\ }%
            \edef\@citeb{\expandafter\@firstofone\@citeb\@empty}%
            \if@filesw\immediate\write\@auxout{\string\citation{\@citeb}}\fi
            \@ifundefined{b@\@citeb}{\hbox{\reset@font\missingcitation}%
                \G@refundefinedtrue
                \@latex@warning
                {Citation `\@citeb' on page \thepage \space undefined}}%
            {\@cite@ofmt{\csname b@\@citeb\endcsname}}}}{#1}}
\makeatother

\newcommand{\gem}[1]{\mbox{\textsc{gem}}}
\newcommand{\abr}[1]{\textsc{#1}}

\newcommand{\hidetext}[1]{}
\newcommand{\ignore}[1]{}

\ifcomment
    \newcommand{\pinaforecomment}[3]{\colorbox{#1}{\parbox{.8\linewidth}{#2: #3}}}

    \newcommand{\prtodo}[1]{\pinaforecomment{lightblue}{pr}{#1}}
    \newcommand{\prtodoi}[1]{\pinaforecomment{lightblue}{pr}{#1}}
\else
    \newcommand{\pinaforecomment}[3]{}
    \newcommand{\prtodo}[1]{}
    \newcommand{\prtodoi}[1]{}
\fi

\newcommand{\smallurl}[1]{ \begin{tiny}\url{#1}\end{tiny}}

\definecolor{lightblue}{HTML}{3cc7ea}
\definecolor{CUgold}{HTML}{CFB87C}
\definecolor{grey}{rgb}{0.95,0.95,0.95}
\definecolor{ceil}{rgb}{0.57, 0.63, 0.81}
\definecolor{UMDred}{HTML}{ed1c24}
\definecolor{UMDyellow}{HTML}{ffc20e}

\newcommand{\bluebold}[1]{\textcolor{darkblue}{\large{\textbf{#1}}}}

\usepackage{listings}
\usepackage{float}
\usepackage{caption}

\def\savelastnode{\pgfextra\edef\tmpA{\tikzlastnode}\endpgfextra}
\def\restorelastnode{\pgfextra\edef\tikzlastnode{\tmpA}\endpgfextra}
\tikzstyle{mybox} = [draw=black, fill=yellow!20, thick,
    rectangle, rounded corners, inner sep=10pt, inner ysep=20pt]
\tikzstyle{mynewbox} = [draw=black, fill=yellow!20, thick,
    rectangle, rounded corners, inner sep=5pt, inner ysep=3pt]
\tikzstyle{fancytitle} =[fill=black, text=white]
\tikzstyle{title} = [append after command={%
    \savelastnode node[fancytitle,right=10pt] at (\tikzlastnode.north west)%
    {\large{#1}}\restorelastnode}]

\newcommand{\incidentset}[1]{%
  \ifnum#1=1 
    event set%
  \else
    \ifnum#1=2
        event sets%
    \else
        Event Set%
    \fi
  \fi
}

\newcommand{\llm}{\abr{llm}}
\newcommand{\grpo}{\abr{grpo}}
\newcommand{\rouge}{\abr{rouge}}
\newcommand{\bertscore}{\abr{berts}core}
\newcommand{\sft}{\abr{sft}}
\newcommand{\rl}{\abr{rl}}
\newcommand{\grm}{\abr{grm}}
\newcommand{\grmLlama}{\grm{}-llama-3B}
\newcommand{\preferenceBert}{Pref\abr{bert}}
\newcommand{\el}{\abr{eli5}}
\newcommand{\longform}{LongForm}
\newcommand{\alpaca}{Alpaca}

\newcommand{\rlvr}{\abr{rlvr}}
\newcommand{\lmJudge}{\abr{llm}-as-a-judge}

\usepackage{times}
\usepackage{latexsym}
\usepackage[T1]{fontenc}
\usepackage[utf8]{inputenc}
\usepackage{microtype}
\usepackage{inconsolata}
\usepackage{graphicx}
\usepackage{booktabs}
\usepackage{amsmath}

\usepackage{algorithm}
\usepackage{algpseudocode}

\newcommand{\wxy}[1]{\textbf{\textcolor{blue}{$\leftarrow$ #1}}}

\title{
Semantically-Aware Rewards for Open-Ended R1 Training in Free-Form Generation}

\author{Zongxia Li \hspace{0.3cm} Yapei Chang \hspace{0.3cm} Yuhang Zhou \hspace{0.3cm} Xiyang Wu \hspace{0.3cm} 
 Zichao Liang  \\  \textbf{Yoo Yeon Sung}   \hspace{0.3cm} \textbf{Jordan Lee Boyd-Graber} \\
University of Maryland, College Park \\
  \texttt{\{zli12321, yapeic, yuhang, wuxiyang, zliang14, yysung53, jbg\}@umd.edu}\\
}

\begin{document}
\maketitle

\begin{abstract} 
Evaluating open-ended long-form generation is challenging because it is hard to define what clearly separates good from bad outputs. Existing methods often miss key aspects like coherence, style, or relevance, or are biased by pretraining data, making open-ended long-form evaluation an underexplored problem.
To address this gap, we propose \preferenceBert{}, a scoring model for evaluating open-ended long-form generation in \grpo{} and guiding its training with distinct rewards for good and bad outputs.
Trained on two response evaluation datasets with diverse long-form styles and Likert-rated quality, \preferenceBert{} effectively supports \grpo{} by offering better semantic reward feedback than traditional metrics \rouge{}-L and \bertscore{} do.
Through comprehensive evaluations,  including \lmJudge{}, human ratings, and qualitative analysis, we show that \preferenceBert{}, trained on multi-sentence and paragraph-length responses, remains reliable across varied long passages and aligns well with the verifiable rewards \grpo{} needs.
Human evaluations confirm that using \preferenceBert{} as the reward signal  to train policy models yields responses better aligned with human preferences than those trained with traditional metrics.
Our code is available at \url{https://github.com/zli12321/long_form_rl}.

\end{abstract}

\section{Introduction}
\label{sec:intro}

\begin{figure*}[htbp]
    \hspace*{-0.3cm} % Adjust this value to control left shift
    \centering
    \includegraphics[width=0.8\textwidth]{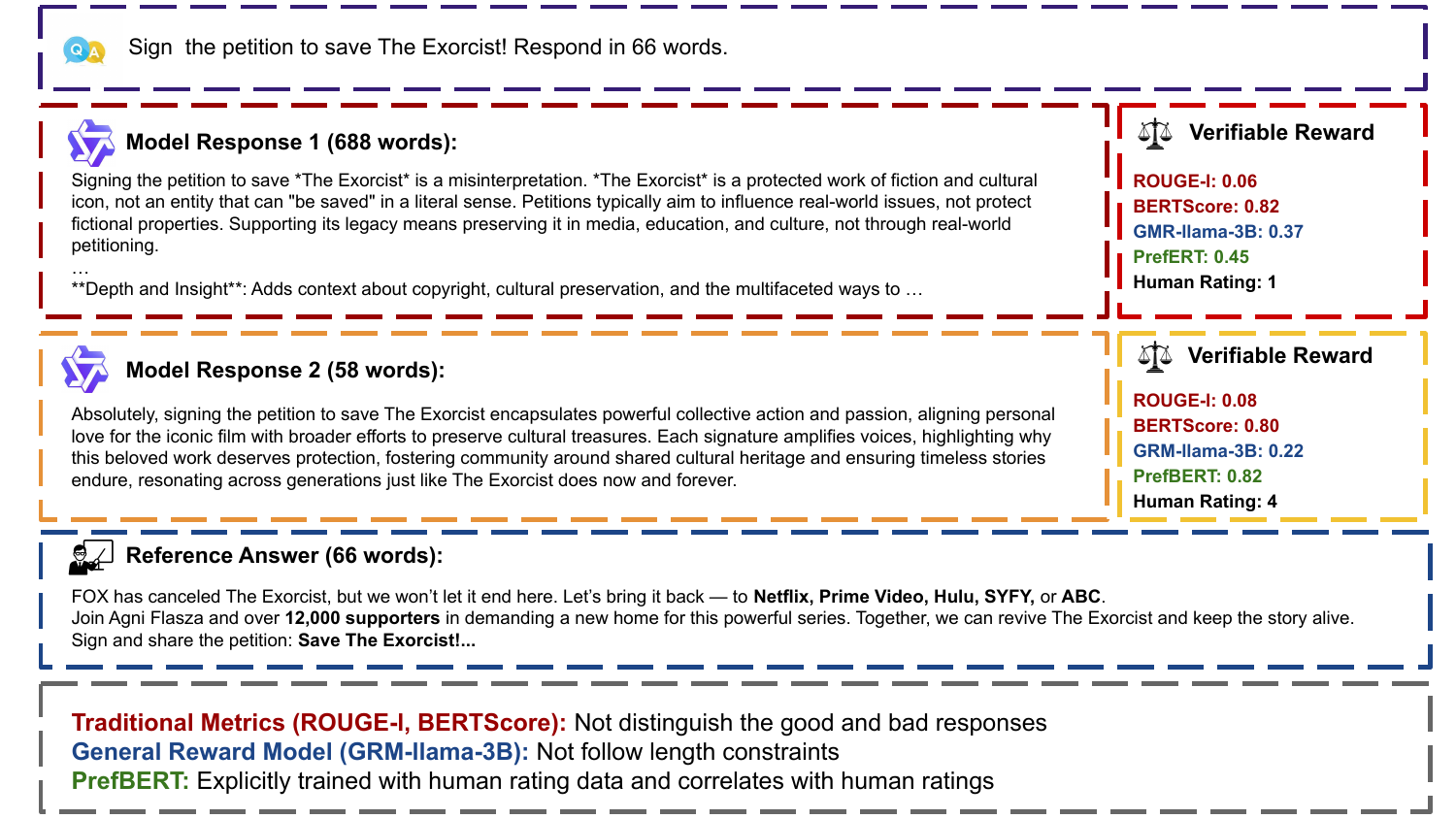}
    \caption{
    % Traditional verifiable rewards do not distinguish between good and bad responses, while \preferenceBert{} trained with human Likert rating data is more correlated with human judgments, which can serve as strong verifiable reward signals for \grpo{}.
    Our proposed reward model \preferenceBert{} better reflects human judgments by distinguishing the good responses from the bad ones than other traditional metrics, generalized reward model.
    }
    % \jbgcomment{The teaswer doesn't make it clear what you do.  New data?  New model?  New eval?}
    \label{fig:teaser}
    \vspace{-10pt}
\end{figure*}

Identifying the good and bad generations is the key to the success of \rlvr{} to improve \llm{}s' abilities on structured rule-based tasks such as mathematical problem-solving~\cite{ahn2024large}, classification~\cite{rouzegar2024enhancing}, and game planning~\cite{shao2019surveydeepreinforcementlearning}. 
%
% Reinforcement learning with verifiable rewards (\rlvr{}) has become an effective method for aligning \llm{}s with human preferences, demonstrating remarkable success on structured rule-based tasks such as mathematical problem-solving, classification, and game planning~\cite{shao2019surveydeepreinforcementlearning}. 
%
\rl{} algorithms like Group Relative Policy Optimization (\grpo{}) excel in mathematical problem-solving tasks by leveraging clear, rule-based reward signals (e.g., correctness) that effectively distinguish between correct and incorrect responses~\cite{deepseekai2025deepseekr1incentivizingreasoningcapability}.
However, despite these advances, extending \grpo{} to open-ended, long-form text generation tasks, such as writing a poem, creative compositions, instruction-following, remains largely underexplored.
A key challenge in applying \grpo{} to open-ended generation is the lack of objective evaluation criteria, making it difficult to distinguish high-quality responses from poor ones.~\cite{krishna2021hurdlesprogresslongformquestion}. 
%
% \zichao{Unlike structured tasks where success is binary or rule-based, long-form generation requires models to balance coherence, fluency, informativeness, and creativity which are all subjective dimensions that are hard to define with discrete rules. This makes reward design a fundamental challenge: how do we guide a model to write better, not just longer? A key barrier is the lack of reliable, scalable evaluation methods. Traditional metrics like \rouge{}~\cite{lin-04} and \bertscore{}~\cite{zhang2020bertscoreevaluatingtextgeneration}—which measure lexical or embedding-level similarity to a reference—often fail to capture qualities that humans actually care about, such as clarity, relevance, and stylistic appropriateness~\cite{chen-etal-2019-evaluating, li2024pedantscheapeffectiveinterpretable}. As a result, these metrics correlate poorly with human preferences, making them unreliable reward signals in reinforcement learning pipelines.}. 
Unlike structured tasks evaluated by explicit correctness criteria, long-form generation requires models to balance coherence, fluency, and informativeness which are all subjective dimensions that are hard to define with discrete rules. 
This makes reward design a fundamental challenge in \grpo{}: How do we guide a model to write better, not just longer? 

A key barrier is the lack of reliable, scalable evaluation methods. Traditional metrics like \rouge{}~\cite{lin-04} and \bertscore{}~\cite{zhang2020bertscoreevaluatingtextgeneration}—which measure lexical or embedding-level similarity to a reference—often fail to capture qualities that humans actually care about, such as clarity, relevance, and stylistic appropriateness~\cite{chen-etal-2019-evaluating, li2024pedantscheapeffectiveinterpretable}. 
Thus, traditional metrics correlate poorly with human preferences, making them unreliable reward signals in \grpo{} pipelines. 
Recent work has applied general reward models (\grm{})~\cite{lambert2024rewardbenchevaluatingrewardmodels} in \abr{rlhf} to select preferred responses, but they are not designed as verifiable rewards for \grpo{}, and their large size introduces latency and high computational cost.
As shown in Figure~\ref{fig:teaser}, traditional metrics fail to distinguish between clearly better and worse responses, assigning similar scores to both, while \grm{} rewards often correlate with length rather than content quality.

% As illustrated in Figure~\ref{fig:teaser}, traditional metrics do not distinguish the good and bad generations and give similar scores for the two responses while one is clearly better than the other, while \grm{} rewards based on length instead of the quality of the actual content.

% models trained with large general reward models (\grm{}) may produce verbose responses that violate clear task instructions- such as exceeding a word limit- yet still receive high rewards, highlighting the danger of length-biased reward signals in open-ended generation.
%

% \textcolor{red}{discuss grm deparately. Add some other disadvantages of grm, latency, resource, cost, not specifically trained for rewards.}

% Unlike structured tasks evaluated by explicit correctness criteria, open-ended tasks do not have clearly defined correct answers, and metrics such as \rouge{}~\cite{lin-04} and \bertscore{}~\cite{zhang2020bertscoreevaluatingtextgeneration}—typically used to evaluate responses against a reference—often rely heavily on lexical or embedding-based semantic similarity. 
% %
% These traditional metrics generally fail to capture  quality differences that align closely with human judgments in the semantic level, resulting in evaluations that correlate poorly with human preferences~\cite{chen-etal-2019-evaluating, li2024pedantscheapeffectiveinterpretable}.

To address open-ended evaluation in \grpo{}, we propose \preferenceBert{}—a lightweight Modern\abr{bert}~\cite{warner2024smarterbetterfasterlonger} model trained on diverse long-form responses and human ratings to provide fine-grained quality evaluation.
% To overcome open-ended evaluation challenge in \grpo{}, we propose using a lightweight Modern\abr{bert}~\cite{warner2024smarterbetterfasterlonger} to train a reward model (\preferenceBert{}) explicitly on diverse long-form answer quality evaluation and human rating data to provide a more fine-grained evaluation on long-form responses.
% , or available off-the-shelf general reward models (\grm{})~\cite{lambert2024rewardbenchevaluatingrewardmodels} trained on human preference data. 
%
% We hypothesize that preference-based reward models, which can capture human preferences, response quality, and semantic meaning, provide more robust and effective reward signals for \grpo{} to distinguish the good and bad responses than traditional evaluation metrics.
We hypothesize that \preferenceBert{}, by capturing human preferences, quality, and semantics, provides stronger reward signals than traditional metrics, enabling \grpo{} to train models that generate higher-quality responses.
% \wxy{add more content here}

%
% \zichao{Notably, we observe that large off-the-shelf reward models such as \grm{} often reward longer, more verbose responses, even when prompts request concise answers or specific word counts. As shown in (Appendix.Figure~\ref{fig:qual_exorcist}), this length bias results in reward inflation for outputs that violate task instructions, highlighting the need for more instruction-following and scale-sensitive reward models like \preferenceBert{}.}

%
% \wxy{Again, the background info can be greatly shrinked, and the intro can be shorted into 2 para}

Through extensive evaluations—including \lmJudge{} point-wise scoring, pairwise Bradley-Terry ranking analyses~\cite{bradly-terry}, and human rating and qualitative evaluation—we show that leveraging stronger, \preferenceBert{} significantly enhances the quality of open-ended text generation across three long-form datasets-- \el{}~\cite{fan2019eli5longformquestion}, \alpaca{}~\cite{alpaca}, \longform{}~\cite{koksal2023longform}. 
Our results show that using improved fine-grained evaluators as reward signals in \grpo{} leads to better alignment with human preferences for open-ended response generation compared to traditional metric-based rewardsand.
Furthermore, smaller models (e.g., Qwen-2.5-3B-Instruct~\cite{qwen2025qwen25technicalreport}) trained with our enhanced reward models generate similarly preferred and concise responses as their larger counterparts (e.g., Qwen-2.5-32/72B-Instruct), and outperform models trained with traditional supervised fine-tuning (\abr{sft}) in preference quality.
Our contributions are:
\begin{itemize}
    \item We introduce \preferenceBert{}, a lightweight long-form reward model that can be easily extended to \grpo{} training.
    % an efficient fine-tuned \grm{} as robust evaluators for open-ended long-form text generation within \grpo{} training frameworks, reducing the need for large-scale human preference annotation.
    \item We validate using \preferenceBert{} in \grpo{} to train models on across multiple open-ended generation benchmarks (\el{}, \alpaca{}, \longform{}), showing resulting model have an overall higher alignment with human preferences compared to traditional metrics and \sft{} training.
    \item Through human expert annotations, we further confirm that models trained with \preferenceBert{} align better with human preferences than traditional metrics as rewards, showing a promising direction for using \grpo{} to improve open-ended generation.
\end{itemize}

\section{Related Work}
\label{sec:related}

\paragraph{\rl{} for \llm{} alignment:}
\rl{} is pivotal in aligning \llm{}s with human preferences by optimizing non-differentiable objectives, making it valuable for tasks like dialogue~\cite{li2016deepreinforcementlearningdialogue}, summarization~\cite{roit2023factuallyconsistentsummarizationreinforcement}, and code generation~\cite{le2022coderlmasteringcodegeneration}.
Popular \abr{rlhf} methods include \abr{dpo}, which applies a classification loss over preference data, and \abr{ppo}, which trains a reward model to guide generation~\cite{wu2023pairwiseproximalpolicyoptimization}. However, both of them require substantial human-annotated data or computational resources.
To address this, \grpo{}~\cite{deepseekai2025deepseekr1incentivizingreasoningcapability} leverages self-generated data and simple, verifiable reward functions to reduce annotation needs, especially for tasks with clear correctness signals like math\cite{liu2025understandingr1zeroliketrainingcritical}.
Extensions such as \abr{dapo}\cite{yu2025dapo}, \grpo{}-\abr{lead}\cite{zhang2025grpo} and \abr{DISCO} \cite{zhou2025disco} broaden \grpo{}'s capabilities in math problem solving. However, these approaches still rely on rule-based reward designs, leaving their application to open-ended, long-form generation tasks—where there is not a fixed correct response—underexplored.

\paragraph{Free-form and open-ended evaluation:}
Evaluating long-form and open-ended generation in \llm{}s remains difficult~\cite{krishna2021hurdlesprogresslongformquestion, chen-etal-2019-evaluating}. Unlike short-form tasks with clear correctness signals, long-form outputs, like summaries, dialogues, or open-ended answers, lack binary ground truths and require assessing coherence, factuality, structure, and helpfulness~\cite{chatbotArena, fabbri2021summevalreevaluatingsummarizationevaluation, Li_2025_CVPR}.
Traditional metrics such as \rouge{}\cite{lin-04}, \abr{bleu}\cite{bleu}, and \bertscore{}\cite{zhang2020bertscoreevaluatingtextgeneration} rely on token overlap or embeddings but poorly reflect semantic or pragmatic qualities, often misaligning with human judgments\cite{chen-etal-2019-evaluating}.
To overcome this, \llm{}-as-a-judge offers more nuanced evaluation through pairwise comparisons or Likert ratings, aligning better with human preferences~\cite{chatbotArena, gu2025surveyllmasajudge, zheng2023judgingllmasajudgemtbenchchatbot, zhou2025mergeme}. However, this approach introduces heavy computational costs, especially in \grpo{} where multiple generations and evaluations per prompt are needed~\cite{deepscaler2025}, limiting accessibility for users with modest resources.
Some efforts fine-tune small models using human ratings to act as open-source judges~\cite{kim2024prometheus2opensource, yang2024regularizinghiddenstatesenables, Chen_2020, zhou2024teaching}, useful for ranking or evaluation~\cite{li2024pedantscheapeffectiveinterpretable, krumdick2025freelabelslimitationsllmasajudge}. Still, few explore using them as verifiable rewards in training, leaving this an open area for research.

\section{Conceptual Backgrounds}

% \jbgcomment{You need to make it clearer what is existing and what your contribution is.  You might want to separate this section into to.}

\label{sec:reward_model}
% In this section, we provide background on \grpo{} training and briefly review existing long-form verifiable rewards. Then we explain how \preferenceBert{} is integrated as a reward signal to enable more robust and effective learning.
In this section, we first review \grpo{} training and existing verifiable rewards for long-form generation, and then introduce how \preferenceBert{} can be used as a reward signal to improve learning robustness and effectiveness.

% we provide background on \grpo{} and details on how we integrate rewards for open-ended, long-form evaluation into its training process.
% \yapei{for this intro para, we can take stuff from the start of 2.2. so we briefly establish why we use GRPO before jumping into how it works. can possibly also compare GRPO to prev methods like PPO/DPO and mention why we choose GRPO.}

% \wxy{Organize this section as follows: 1. Preliminary (GRPO), 2. Other Methods (Basslines), with some takeaway notes and discussions, 3. Your method, in detail, highlight your contributions}

\subsection{Preliminaries on \grpo{}}
\label{section:gpo}
% \textcolor{red}{briefly discuss grpo here}
\grpo{} is an \rl{} algorithm designed to refine language model policies, $\pi_{\phi}$, using reward signals contextualized within a group of candidate responses.
Given a prompt $x$ from dataset $\mathcal{D}$, \grpo{} samples $G$ responses ${y_i} = {y_1, \dots, y_G}$ from the old policy $\pi_{\phi_{\text{old}}}(y|x)$. Each response $y_i$ is assigned a scalar reward $r(x, y_i)$ (detailed in the following subsections). The group-normalized advantage $A(x, y_i)$ is then computed as:
%
% \grpo{} is an \rl{} algorithm designed to refine language model policies, $\pi_{\phi}$, by learning from reward signals that are contextualized within a group of candidate responses. 
% % Unlike methods that rely solely on individual response scores or direct pairwise comparisons, \grpo{} emphasizes the relative performance of a policy's outputs when evaluated as part of a group.
% Specifically, for a given prompt $x$ from dataset $\mathcal{D}$, a total of $G$ responses $\{y_i\} = \{y_1, \dots, y_G\}$ are sampled from the old policy model $\pi_{\phi_{\text{old}}}(y|x)$. Each response $y_i$ receives a scalar signal reward $r(x,y_i)$ (introduced in the following subsections). The group-normalized advantage $A(x,y_i)$ for each response $y_i$ is then calculated as:
\begin{equation}
\label{eq:grpo_advantage}
A(x,y_i) = \frac{r(x,y_i) - \bar{r}(x)}{\sigma_r(x)},
\end{equation}
where $\bar{r}(x) = \frac{1}{G} \sum_{j=1}^{G} r(x,y_j)$ and $\sigma_r(x)$ are the mean and standard deviation, respectively, of rewards $r(x,y_j)$ within the group $Y$. This normalization contextualizes each advantage relative to the group's current performance.

The new policy $\pi_{\phi}(y|x)$ is optimized by maximizing the \grpo{} objective, which combines a clipped surrogate loss with a Kullback-Leibler (KL) divergence penalty \cite{kullback1951information} against a reference model $\pi_{\text{ref}}(y|x)$ for regularization
(Equation~\ref{eq:grpo_objective_detailed}, in Appendix~\ref{app:tech}).

Although originally applied to tasks with explicit, rule-based rewards (e.g., correctness or win/loss), \grpo{}’s reliance on advantage estimation and KL-regularized updates allows it to learn from scalar feedback, making it well-suited for open-ended tasks where response quality lies on a spectrum rather than binary correctness.

\subsection{Existing Methods for Scoring Open-ended Generation}

Current scoring methods for open-ended generation mainly fall into two categories. The first are reference-based metrics, a method commonly used in natural language generation. These methods score the generations over metrics like string overlap or embedding similarity. While easy to apply, they correlate poorly with human preferences on long-form outputs~\citep{chen-etal-2019-evaluating, Chen_2020, kim2024prometheus2opensource, li2024pedantscheapeffectiveinterpretable, gu2025surveyllmasajudge, li2025videohalluevaluatingmitigatingmultimodal}.
The alternative is to use \llm{}s as evaluators, which compute scores via pairwise preference prompts, Likert-scale tuning~\cite{kim2024prometheus2opensource}, or general reward models (\grm{}s) trained on preference data~\cite{lambert2024rewardbenchevaluatingrewardmodels}. While more adaptive to human preferences, they may still inherit biases from their training data.
In our work, we adopt two reference-based metrics, \rouge{} and \bertscore{}, and one reward model–based method, \grm{}-LLaMA-3B, as baselines.

\paragraph{\rouge{}~\citep{lin-04}} is a reference-based metric that measures $n$-gram overlap between generated and reference texts. Variants include \rouge{}-1, \rouge{}-2, and \rouge{}-L, which capture unigram/bigram matches and the longest common subsequence, respectively.

%
% \rouge{} was originally shown to correlate with human preferences~\cite{lin-04} in settings where the goal was to generate responses closely matching a reference answer. 
% %
% However, as \llm{}s now produce human-level outputs, traditional \rouge{} fails to capture diversity, creativity, and quality, making them less consistent with modern human evaluation requirements.

% \paragraph{\bertscore{} \citep{zhang2020bertscoreevaluatingtextgeneration}} is a metric that measures the embedding semantic similarity between the reference answer and the generated answer~\cite{zhang2020bertscoreevaluatingtextgeneration}. 
% %
% \bertscore{} is shown to be more correlated with human judgments than token overlapping methods (\rouge{}) on long-form text generation and translation tasks, but the reliability of both metrics does not hold on modern datasets and models~\cite{bhandari-etal-2020-evaluating}.

\paragraph{\bertscore{}~\citep{zhang2020bertscoreevaluatingtextgeneration}} is a reference-based metric that measures semantic similarity between the reference and generation using contextual embeddings.
It has shown a stronger correlation with human judgments than token overlap metrics like \rouge{} on long-form generation and translation tasks. However, its reliability diminishes on modern datasets and models~\cite{bhandari-etal-2020-evaluating}.

% \paragraph{Generalized Reward Model (\grm{}-llama-3B)} is a reference-free \llm{}-based reward model~\cite{yang2024regularizinghiddenstatesenables} that is finetuned on the decontaminated Skywork preference dataset~\cite{liu2024skywork}, where each data point $x$ contains an input prompt, a chosen response ($y_c$), and a rejected response ($y_r$).
% %
% The reward model is trained to assign higher scores to preferred outputs by minimizing the Bradley-Terry style loss~\cite{bradly-terry} (Equation~\ref{eq:grmBtLoss}).

\paragraph{Generalized Reward Model (\grm{}-LLaMA-3B)} is a reference-free, \llm{}-based reward model~\cite{yang2024regularizinghiddenstatesenables}, fine-tuned on the decontaminated Skywork preference dataset~\cite{liu2024skywork}. Each data point $x$ includes an input prompt, a preferred response ($y_c$), and a rejected response ($y_r$).
The model is trained to favor preferred outputs by minimizing a Bradley-Terry-style loss~\cite{bradly-terry} (Equation~\ref{eq:grmBtLoss}, Appendix~\ref{app:tech}).

%
% \grm{} is commonly used is \abr{rlhf} to fine-tune \llm{}s through preference-based optimization rather than relying on explicit supervision.

\subsection{\preferenceBert{}: Training a Lightweight BERT-Based Reward Model}
\label{sec:preferenceBert}
% \wxy{Need to significantly refine or organize, like what's your structure or training setting of this method, how do you fix previous flaws in this method?}
% While these methods improve evaluation quality, they are computationally expensive and difficult to scale due to the size of the underlying models.
While traditional metrics often fail to capture human preferences, reward models can be computationally expensive to scale due to the size of the underlying models.
To balance quality and efficiency, we propose \preferenceBert{}, a lightweight reference-based model with only 150M parameters, to offer fine-grained reward signals while maintaining training efficiency.
% —serving as a lightweight yet effective alternative to \llm{}-based evaluators in \grpo{} training
%
% In the rest of the section, we discuss reward models and reward metrics in the \grpo{} process. 
% Inspired by ~\citet{kim2024prometheus2opensource, li2024pedantscheapeffectiveinterpretable, Chen_2020}, instead of retraining a large \llm{} as a reference-free reward model on pairwise preference data, we train a reference-based small model with point-wise evaluation.

\paragraph{Creating training data for \preferenceBert{}:}
Each training example has a reference answer, a generated answer, and a Likert score from 1-5 that rates the quality of the generated answer against the reference. 
To ensure balanced quality ratings across both long and short free-form responses, we incorporate training data from the Prometheus-preference~\cite{kim2024prometheus2opensource} and MOCHA~\cite{Chen_2020}.\footnote{Specifically, Prometheus-preference contains 200K fine-grained Likert preference ratings spanning ten categories of evaluation including e.g. adaptive communication, emotional intelligence; the data is primarily long free-form answers where each answer is above 150 tokens. MOCHA contains mid to long length answer evaluation data to judge the overall correctness of the generated response.}
We combine the two datasets and split them into 80\% for training and 20\% for testing. The resulting training set contains 19K examples---substantially smaller than the 80K examples used to train \grmLlama{}.
% These three datasets are used to train the reward model in \grpo{}, not the policy model.
%

% \textcolor{red}{Specifically say this is to train the verifiable reward model, and later datasets are used to train the policy}
%
\paragraph{Training \preferenceBert{}:}
% We train ModernBert \cite{warner2024smarterbetterfasterlonger} (150M parameters) on triplets consisting of a human reference answer, a generated answer, and a corresponding Likert score. The input is structured as \textit{reference answer [SEP] generated answer}. The model output is passed through a linear layer to produce a scalar, followed by a sigmoid to yield a normalized prediction. The target score is scaled to the [0, 1] range as:
% \begin{equation}
% r_i = \frac{s_i - 1}{4}, \quad s_i \in \{1, 2, 3, 4, 5\}.
% \end{equation}
% %
% % We concatenate reference and candidate with a [SEP] token and passed through the model with an additional linear layer to produce a scalar, followed by a sigmoid to yield the predicted normalized score.  All parameters \(\theta\) are learned by minimizing the mean‐squared error loss:
% % \begin{equation}
% % \mathcal{L}(\theta) = \frac{1}{N}\sum_{i=1}^N \bigl(\hat\ell_i - \ell_i\bigr)^2
% % \end{equation}
% %
% where $r_i$ is taken as the reward signal of \grpo{}.

We train ModernBERT~\cite{warner2024smarterbetterfasterlonger} on triplets $(x^{\mathrm{ref}}_i, x^{\mathrm{gen}}_i, s_i)$ where $s_i\in\{1,\dots,5\}$.  We first normalize each gold Likert score to  
\[
r_i = \frac{s_i - 1}{4}\;\in[0,1],
\]  
where $s_i$ is the gold Likert scale, $r_i$ is the normalized Likert score on the $[0,1]$ scale, $x^{ref}$ is the reference answer, and $x^{gen}$ is the generated response. Thus, given $x^{ref}$ and $x^{gen}$, we concatenate them as a single string:
\begin{equation}
x^{\mathrm{pair}} = \texttt{[CLS]}~x^{\mathrm{ref}}~\texttt{[SEP]}~x^{\mathrm{gen}},
\label{eq:paired_input}
\end{equation}
where $x^{\mathrm{pair}}$ is the input string feeds into ModernBERT. Let $\mathbf{h}_i\in\mathbb{R}^d$ be the pooled ModernBERT embedding of $x^{\mathrm{pair}}$.
A linear regressor plus sigmoid yields a prediction  
\begin{equation}
\hat r_i = \sigma\bigl(\mathbf{w}^\top\mathbf{h}_i + b\bigr),
\end{equation}
where $\mathbf{w}\in\mathbb{R}^d$ and $b\in\mathbb{R}$ are the regressor weights and bias, $\sigma(z)=1/(1+e^{-z})$ is the sigmoid activation. $\hat r_i\in(0,1)$ is the predicted normalized score, and is taken as the reward signal of \grpo{}.
Training minimizes the mean-squared error  
\begin{equation}
\mathcal{L}
=\frac{1}{N}\sum_{i=1}^N\bigl(\hat r_i - r_i\bigr)^2,
\end{equation}
where  $\mathcal{L}$ is the batch-averaged MSE loss,  
and $N$ is the number of examples in the batch.

\section{Experiment Setup}
\label{sec:experiment}
% We propose \preferenceBert{} and want to study whether more fine-grained reward signals can improve an \llm{}'s open-ended generation quality than traditional less-fine-grained metrics as reward.
% %
% We will need to conduct experiments on datasets with long-form responses, unlike the short-form QA datasets Trivia~\cite{joshi2017triviaqalargescaledistantly} or math datasets GSM8K~\cite{cobbe2021trainingverifierssolvemath} that have definite answers of correctness, we need to use datasets that have long responses that cover broad spectrum of topics.
%
% We propose \preferenceBert{} to investigate whether stronger reward signals can enhance an \llm{}'s ability in open-ended generation tasks compared to traditional metrics.
With the background on \grpo{} and the various reward signals established, we now outline our experimental setup, including the datasets, base models, and training methods.
%

% In this section, we summarize the datasets and tasks and models we are using, then provide our training groups and automatic evaluation setup. \yapei{can make more specific, e.g., ``In this section, we provide details on our experimental setup, including three datasets covering diverse domains, two base models, and 4 (or 5?) training methods. \jbgcomment{Yes, and explain why you're doing it.  What is the goal?}}

\subsection{Free-Form and Open-Ended Datasets}

% as opposed to short-form QA datasets like TriviaQA~\cite{joshi2017triviaqalargescaledistantly} or GSM8K~\cite{cobbe2021trainingverifierssolvemath}, which focus on questions with definitive correct answers.
%
We use three datasets featuring long-form responses that span a broad spectrum of topics.
They cover diverse styles of long-form responses averaging 185 words, requiring evaluation across dimensions like coherence, conciseness, and informativeness rather than correctness.\footnote{Examples in \S Table~\ref{tab:dataset}.} More details on each dataset are in \S\ref{app:dataset_details}.
% Specifically, we select \el{}\cite{fan2019eli5longformquestion}, \alpaca{}\cite{alpaca2023}, and \longform{}, which represent diverse styles of long-form responses averaging 185 words, requiring evaluation across dimensions like coherence, conciseness, and informativeness rather than correctness.\footnote{Examples in Appendix Table~\ref{tab:dataset}.}

% \wxy{Cut off this part greatly, also give some insights over why you choose these datasets if possible}
\paragraph{\el{} \cite{fan2019eli5longformquestion}} is a collection of questions and answers from Reddit's r/explainlikeimfive community.\footnote{\url{https://www.reddit.com/r/explainlikeimfive/}} 
We sample 10,444 questions as the train set and 1,056 as the test set. 

\paragraph{\alpaca{} \cite{alpaca2023}}
is a collection of 52K instruction-response pairs generated by OpenAI's text-davinci-003 in the style of Self-Instruct~\cite{selfinstruct}. We use 10,444 examples as the train set and 1,334 as the test set.
% is a collection of 52K instruction-response pairs generated by OpenAI's text-davinci-003 to fine-tune LLaMA 7B~\cite{alpaca2023}.\footnote{\url{openai.com}}
% %
% It diverse prompts and corresponding long responses in the style of the Self-Instruct~\cite{selfinstruct}. 
% %
% We use 10,444 examples as train set and 1,334 examples as the test set.

% This dataset enabled the development of Alpaca, a lightweight, instruction-following model that performs comparably to larger models like text-davinci-003, while being more accessible and cost-effective for academic research .

\paragraph{\longform{} \cite{koksal2023longform}}
is built from English documents (e.g., Wikipedia~\cite{wikipedia}, C4~\cite{c4}) paired with reverse-instruction prompts generated by \llm{}s. We exclude coding tasks, sampling 8,648 training and 956 test examples.

We merge the three sampled datasets together as our free-form train/test set.
% We organize them in the order of \alpaca{}, \longform{}, and \el{} to facilitate a curriculum where the model first learns the easy questions, then the hard questions.
% is created by leveraging English corpus examples with reverse instructions.
% %
% It contains diverse set of human-written documents from e.g. Wikipedia~\cite{wikipedia}, C4~\cite{c4}, Stack Exchange~\cite{stackexchange2025}, Big Bench~\cite{bigbench} and the instructions are generated via \llm{}s with task examples spanning from question answering, email writing, story/poem generation, and text summarization.
% %
% We remove the examples that requires coding from our examples since it is considered out of scope and we sample 8,648 questions as the training set and 956 as the test set.
% %
%
% Additionally, we organize the data in the order of \alpaca{}, \longform{}, and \el{} to facilitate a curriculum learning style where the model first learns the easy questions,, then the hard questions. 

\subsection{Training Setup}
% \paragraph{Training \grpo{} on open-ended tasks:} given the combined training datasets, we use reward methods in Section~\ref{sec:reward_model} as rewards to train models using \grpo{} in the OpenRLHF framework~\cite{hu2024openrlhf}: \rouge{}-l, \bertscore{}, \grmLlama{}, \preferenceBert{}.
% %
% We use scores directly from each metric as reward signals of \grpo{}.
% %
% We use Qwen-2.5-Instruct size 1.5B and 3B~\cite{qwen2025qwen25technicalreport} as our base models due to limited computing resources.
% %
% Specifically, for each question or input instruction, we ask the model to directly generate an open-ended response without using chain-of-thought reasoning~\cite{wei2023chainofthoughtpromptingelicitsreasoning} since open-ended questions and instructions are not evaluated based on the traditional correctness, but also based on the overall fluency and informativeness of the answers as a whole.\footnote{Training prompt template in Appendix Table~\ref{tab:training_prompt_temps}}
% %
% Our reference-based reward functions (\rouge{}-l, \bertscore{} and \preferenceBert{}) compute a score by directly comparing each generated response against its corresponding reference response, where reference-free reward model \grmLlama{} simply takes the input prompt and the generated response to compute reward directly.
% %
% We train the model without data shuffling for one epoch, batch size of 4.\footnote{Our computre resource is four A6000 GPUs.}
% %

\paragraph{\grpo{} for open-ended generation:}
We train policy models using \grpo{} within the OpenRLHF framework~\cite{hu2024openrlhf}, optimizing each of the four reward signals from Section~\ref{sec:reward_model} separately: \rouge{}-L, \bertscore{}, \grmLlama{}, and \preferenceBert{}.
% Each score is used directly as the reward signal.
We use two base models, Qwen2.5-1.5B-Instruct and Qwen2.5-3B-Instruct \cite{qwen2025qwen25technicalreport}.
In the training prompt, we encourage models to be factual, relevant, clear, concise, and complete (more details in Appendix Table~\ref{tab:training_prompt_temps}).\footnote{All \grpo{} models are trained on 4 A6000 GPUs for one epoch, with a global batch size of 128, group size of 4, and learning rate of 1e-6. We set both max prompt length and max generation length to 1024.}
% For each input, models generate open-ended responses without chain-of-thought reasoning \cite{wei2023chainofthoughtpromptingelicitsreasoning}, as our evaluation emphasizes conciseness and informativeness, not correctness.\footnote{Training prompt templates are shown in Appendix Table~\ref{tab:training_prompt_temps}.}
%
% Reference-based reward functions (\rouge{}-L, \bertscore{}, and \preferenceBert{}) are computed by comparing generated response against gold references, while the reference-free \grmLlama{} scores responses based solely on the prompt and generation.
%
% and no data shuffling 

\paragraph{Supervised Finetuning (\sft{}):} Using the reference responses as ground truth, we run \sft{} on Qwen2.5-Instruct size 1.5B and 3B.\footnote{All \sft{} models are trained on 4 A6000 GPUs for three epochs, with with a global batch size of 128, learning rate of 1e-5, and max tokens of 4096.}
% \wxy{Add some details here} 

\begin{table*}[ht]
  \tiny
  \centering
\resizebox{\textwidth}{!}{%
\begin{tabular}{%
      >{\raggedright\arraybackslash}p{2.5cm} 
      rrrr|rrrr|rrrr
    }
\toprule
& \multicolumn{4}{c|}{\bfseries Mean Likert Scores} 
& \multicolumn{4}{c|}{\bfseries Success Rates with Score $\geq4$ (\%)} 
& \multicolumn{4}{c}{\bfseries Bradley–Terry Win Rate (\%)} \\
\cmidrule(lr){2-5} \cmidrule(lr){6-9} \cmidrule(lr){10-13}
\textbf{Model} 
  & \el{} & \longform{} & \alpaca{} & Overall 
  & \el{} & \longform{} & \alpaca{} & Overall 
  & \el{} & \longform{} & \alpaca{} & Overall \\
\midrule
\multicolumn{13}{c}{Base \llm{}} \\
\hline
Qwen2.5-72B-Instruct  & 4.39 & 3.56 & 4.23 & 4.13 & 96.59 & 38.91 & 87.93 & 79.25 & 21.94 & 19.87 & 16.11 & 19.33 \\   
% & 4.13 & 3.12 & 3.88 & 3.79 & 86.63 & 22.38 & 72.41 & 65.73 &  8.93 &  6.65 &  6.74 &  7.64 \\
Qwen2.5-32B-Instruct     & 4.28 & 3.52 & 4.29 & 4.10 & 94.15 & 37.55 & 89.51 & 78.47 & 15.43 & 18.24 & 20.68 & 18.09 \\
% & 4.10 & 2.96 & 3.89 & 3.74 & 87.02 & 18.62 & 73.61 & 65.37 &  8.01 &  4.81 &  6.59 &  6.72 \\
Qwen2.5-7B-Instruct      & 4.04 & 2.95 & 3.82 & 3.69 & 77.38 & 18.83 & 66.64 & 59.10 &  7.54 &  4.94 &  5.95 &  6.37 \\
Qwen2.5-3B-Instruct      & 3.90 & 2.88 & 3.75 & 3.59 & 66.00 & 17.99 & 63.57 & 53.22 &  5.18 &  4.17 &  5.26 &  5.09 \\
Qwen2.5-1.5B-Instruct    & 3.61 & 2.26 & 3.44 & 3.21 & 49.16 & 10.25 & 47.38 & 38.87 &  2.69 &  1.49 &  2.89 &  2.51 \\
\midrule
\multicolumn{13}{c}{RL-Finetuned Policy Models (\grpo{})} \\
\hline
3B-\grmLlama{}           & 4.29 & 3.37 & 4.07 & 3.99 & 91.97 & 30.02 & 77.51 & 71.55 & 15.89 & 12.59 & 11.52 & 13.53 \\
3B-\preferenceBert{}      & 4.09 & 3.54 & 4.16 & \cellcolor{cyan!20}3.98 & 84.64 & 28.87 & 84.41 & \cellcolor{cyan!20}70.70 &  7.92 & 18.00 & 12.54 & \cellcolor{cyan!20}11.56 \\
% 3B-random          & 3.83 & 2.68 & 3.62 & 3.47 & 65.30 & 9.10 & 59.60 & 49.35 &  3.63 &  2.56 &  3.43 &  3.37 \\
3B-\bertscore{}          & 3.79 & 2.79 & 3.63 & 3.49 & 60.73 & 11.09 & 59.30 & 47.89 &  3.40 &  3.19 &  3.72 &  3.60 \\
3B-\rouge{}-L            & 3.66 & 2.69 & 3.51 & 3.37 & 51.16 &  7.32 & 52.55 & 40.74 &  2.40 &  2.57 &  2.78 &  2.72 \\
1.5B-\grmLlama{}         & 4.13 & 3.12 & 3.88 & 3.79 & 86.63 & 22.38 & 72.41 & 65.73 &  8.93 &  6.65 &  6.74 &  7.64 \\
1.5B-\preferenceBert{}   & 4.10 & 2.96 & 3.89 & \cellcolor{cyan!20}3.74 & 87.02 & 18.62 & 73.61 & \cellcolor{cyan!20}65.37 &  8.01 &  4.81 &  6.59 &  \cellcolor{cyan!20}6.72 \\
% 1.5B-random   & 3.29 &  2.25 & 3.27 & 3.02 & 28.21  & 4.08 & 35.16 & 24.62 &  1.06 &  1.14 &  1.53 &  1.35 \\
1.5B-\rouge{}-L          & 2.66 & 1.98 & 3.04 & 2.62 &  5.72 &  1.05 & 17.92 &  8.79 &  0.28 &  0.72 &  0.90 &  0.69 \\
1.5B-\bertscore{}        & 2.34 & 1.86 & 3.05 & 2.47 &  0.90 &  0.42 & 17.39 &  6.50 &  0.14 &  0.55 &  0.88 &  0.52 \\
\midrule
\multicolumn{13}{c}{Supervised Finetuning (\sft{})} \\
\hline
3B-sft                   & 2.19 & 2.21 & 3.32 & 2.59 &  2.51 &  1.78 & 36.58 & 14.14 &  0.12 &  1.13 &  1.68 &  0.82 \\
1.5B-sft                 & 2.18 & 2.15 & 3.33 & 2.57 &  2.63 &  1.67 & 37.93 & 14.64 &  0.12 &  1.09 &  1.73 &  0.82 \\
\bottomrule
\end{tabular}
}
  \caption{Evaluation of model outputs via GPT-4 as a judge across different instruction tuning and reward optimization strategies. Groupings show comparisons between SoTA baselines,\rl{}-finetuned models using various reward functions, and supervised finetuning (\sft{}). Larger models are generally stronger, though models fine-tuned with better-aligned reward functions (e.g., \grmLlama{}) may show inflated automatic metrics due to biases like verbosity. 
  % An interesting result is that using random reward achieves higher average score and success rates than using \rouge{}-l and \bertscore{} as rewards, but still falls short behind using semantic evaluators and worse than its base models.
  }
  \label{tab:combined_full_performance}
  \vspace{-10pt}
\end{table*}

\section{Automatic Evaluation}
\label{sec:results}
% \jbgcomment{Too abstract, what responses?}
% \jbgcomment{Likert should always be capitalized}
% \yapei{move the following sentences to sec 4?}
% We use the trained models to generate responses on our test dataset.
%
We use our test set to evaluate our trained models, as well as larger off-the-shelf models: Qwen2.5-Instruct 7B, 32B, and 72B.
For evaluation, we use \llm{}-as-a-judge to evaluate the quality of the responses for different models as they can be strong alternative evaluators of humans~\cite{chiang2023largelanguagemodelsalternative}.
Overall, models trained with our lightweight \preferenceBert{} performs competitively with those trained with the much larger \grmLlama{}, and both substantially outperform models trained with token-overlap metrics or \sft{}. In addition, \preferenceBert{}-trained models at 1.5B and 3B scale rival or exceed the performance of Qwen2.5-7B-Instruct, despite having far fewer parameters.
%
% Specifically, to ensure a more robust automatic evaluation, we use both point-wise Likert scale evaluation and pairwise preference evaluation. 
% %
% Point-wise evaluation is a new era of automatic evaluation that assigns an absolute overall quality score to each response on Likert scale~\cite{fabbri2021summevalreevaluatingsummarizationevaluation}. 
% %
% In contrast, pairwise preference evaluation requires the judge to directly compare two responses and select the better one, yielding more consistent ordinal judgments by avoiding scale-interpretation ambiguities.
% %
% The popular benchmark Chatbot Arena uses pairwise comparison by presenting users with two chatbot responses side-by-side and asking them to vote for their preferred answer, thereby aggregating these direct preferences into model rankings.~\cite{chatbotArena}.
% %
% Point-wise likert scale is easier to compare and rank multiple models at the same time, while pairwise comparison is more complicated and usually requires Bradly-Terry ranking system.

\subsection{Evaluation Metrics}

\paragraph{Point-wise evaluation:}
Point-wise evaluation assigns an absolute overall quality score to each response on a Likert scale~\cite{fabbri2021summevalreevaluatingsummarizationevaluation}.
We use GPT-4 as a judge to first provide some reasoning, then assign a score between 1 to 5 to the generated response, considering aspects like factuality, relevance, clarity and organization, conciseness, and completeness (detailed prompt in Table~\ref{tab:prompt_templates_pointwise_eval}).\footnote{\citet{chiang2023closerlookautomaticevaluation} shows that first analyze the response then give a rating score yields the best correlation with human judgments.}
%
% We use GPT-4 as a judge to first give a Likert score between 1 to 5 for the generated response.\footnote{Evaluation prompt template in \S Table~\ref{app:eval_prompt_template}}
% %
% Specifically, we give GPT-4 the input question/instruction, the reference answer, and the generated response and use chain-of-thought to first give reasoning and analysis of the response then output an overall score based on perspectives of factual consistency, relevance, clarity and organization, conciseness, and completeness (Detailed definitions in \S Table~\ref{tab:prompt_templates_pointwise_eval}).\footnote{\citet{chiang2023closerlookautomaticevaluation} shows that first analyze the response then give a rating score yields the best correlation with human judgments.}
%
We use two metrics—\textit{mean Likert score} (the average overall score) and \textit{success rate} (the percentage of responses that receive a score $\geq 4$)—to evaluate the quality of model responses.

\paragraph{Pairwise preference evaluation:}
From the Likert scores, we derive pairwise comparisons to compute Bradley–Terry win rates. This approach reduces rating noise by focusing on relative preferences rather than absolute scales, which has been shown to yield more reliable comparisons in subjective evaluation settings~\cite{bai2022traininghelpfulharmlessassistant, stiennon2022learningsummarizehumanfeedback}.
For each prompt, we compare the \llm{} ratings between every pair of models. A tie is recorded when both receive the same rating, and a win is assigned to the model with the higher rating.
We use the Bradley-Terry model to compute the probability \textit{win rate} of each model on the three datasets.

\subsection{Results and Discussion}

Table~\ref{tab:combined_full_performance} summarizes model performance across instruction-following tasks using Likert scores, success rates, and Bradley–Terry win rates and \S Figure~\ref{app:reward_curves} shows the training curves of the 3B policy models. Below, we discuss our findings.

% \paragraph{\preferenceBert{} is competitive with \grmLlama{}.}
% Models trained with \preferenceBert{} perform comparably to those trained with \grmLlama{} at both the 3B and 1.5B parameter scales. This is particularly notable given that \preferenceBert{} has 20 times fewer parameters and was trained on only 19K examples, compared to 80K for \grmLlama{}. As discussed later in \S\ref{subsec:qual_analysis}, human evaluations reveal that \grmLlama{} models tend to produce long and overly verbose outputs, which likely inflate their scores in Table \ref{tab:combined_full_performance} due to LLM judges' bias toward longer responses.

% \paragraph{Policy models trained with \preferenceBert{} and \grmLlama{} generate responses that are rated higher by \lmJudges{} .}
\paragraph{Policy models trained with \preferenceBert{} and \grmLlama{} achieve the highest ratings from \lmJudge{} among all evaluated policy models.}
3B-\preferenceBert{} and 3B-\grmLlama{} both have the highest \lmJudge{} average ratings and success rates among all policy models.
However, \preferenceBert{} is more efficient, having 20 times fewer parameters and was trained on only 19K examples, compared to 80K for \grmLlama{}. In addition, as discussed later in Section~\ref{subsec:qual_analysis}, human evaluations reveal that \grmLlama{} models tend to produce long and overly verbose outputs, which likely inflate their scores in Table~\ref{tab:combined_full_performance} due to \llm{} judges' bias toward longer responses.

\paragraph{\preferenceBert{}-trained models rival much larger models.}
Despite their smaller size, \preferenceBert{}-trained models at 1.5B and 3B scale match or exceed the performance of Qwen2.5-7B-Instruct across all metrics. For example, 3B-\preferenceBert{} outperforms Qwen2.5-7B-Instruct in Likert score (3.98 vs. 3.69), success rate (\textbf{70.70\%} vs. 59.10\%), and win rate (\textbf{11.56\%} vs. 6.37\%). While Qwen2.5-32B and 72B are stronger overall, these results highlight how a well-aligned reward model can enable smaller LLMs to compete with much larger ones on open-ended long-form generations.

% \paragraph{Policy models trained with traditional metrics lead to poor alignment.}
% Models trained with standard metrics like \rouge{}-L and \bertscore{} perform the worst across all evaluation dimensions. Notably, 3B-\bertscore{} and 3B-\rouge{} models lag behind 1.5B-\preferenceBert{} by large margins, despite having more parameters.

\paragraph{The pattern that using semantic verifiable reward for training produces better policy models than using traditional metric-based rewards in long-form generation remains consistent across model sizes.}
Models trained with standard metrics, such as 1.5B-\rouge{}-L and 1.5B-\bertscore{}, perform worse across all evaluation dimensions compared to 1.5B-\preferenceBert{} and 1.5B-\grmLlama{}; this trend also holds for the 3B model sizes. 
% . Notably, 3B-\bertscore{} and 3B-\rouge{} models lag behind 1.5B-\preferenceBert{} by large margins, despite having more parameters.

\paragraph{\sft{} underperforms across the board.}
Policy models trained with \sft{} achieve lower scores across all evaluation dimensions than all models trained with \grpo{}, regardless of the reward used, except for 1.5B-\bertscore{}.
Their success and win rates are the lowest across all datasets.
Even the small 1.5B-\preferenceBert{} and 1.5B-\grmLlama{} substantially outperform the 3B-\sft{} model on average Likert score by around 1.1, success rate by 50\%m and Bradley-Terry win rate by 6\%.

\begin{figure*}[tbp]
    % \hspace*{-0.3cm} % Adjust this value to control left shift
    \centering
    \includegraphics[width=1.0\textwidth]{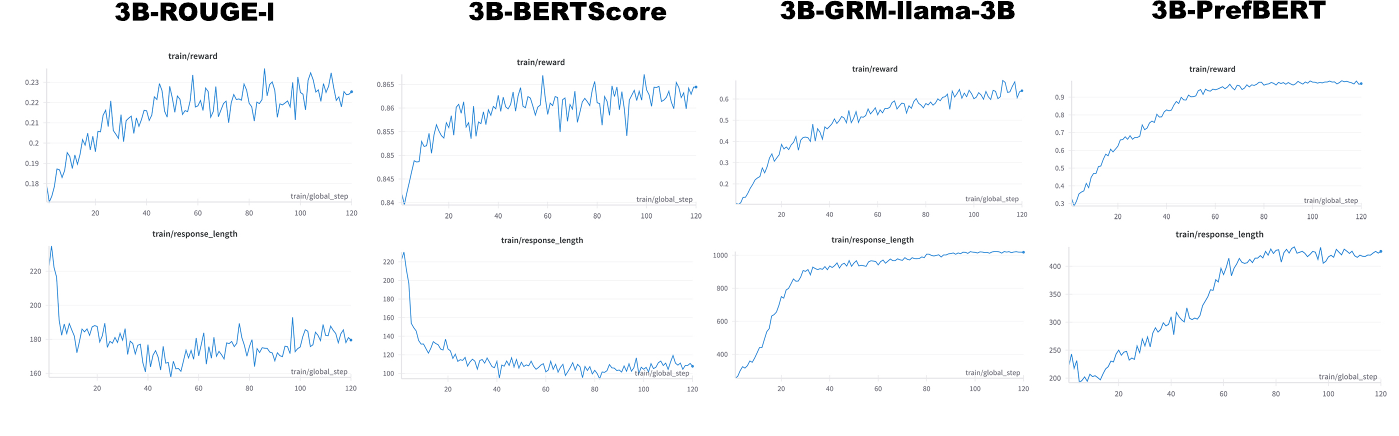}
    \caption{
    % Traditional verifiable rewards do not distinguish between good and bad responses, while \preferenceBert{} trained with human Likert rating data is more correlated with human judgments, which can serve as strong verifiable reward signals for \grpo{}.
    Reward curves during \grpo{} training show key differences across reward functions. Traditional metrics show minimal reward change—only 0.05 for \rouge{}-L and 0.25 for \bertscore{}—indicating limited model improvement across all global training steps. In contrast, 3B-\grmLlama{}'s reward is strongly correlated with response length; by step 60, it already generates the maximum allowed tokens (1,024), causing reward values to plateau around 0.6. \preferenceBert{} shows a more meaningful reward progression, not strictly tied to length, suggesting it favors responses of an optimal length rather than simply longer outputs.
    }
    % \jbgcomment{The teaswer doesn't make it clear what you do.  New data?  New model?  New eval?}
    \label{fig:training_curves}
    \vspace{-10pt}
\end{figure*}

\subsection{Reward Learning Curves} \label{app:reward_curves}
% \textcolor{red}{Discuss the reward curves}
% due to \rouge{}-l' lack of semantic evaluation property, it cannot give meaningful reward feedback signal to the model, and the model-generated responses are quite different from the reference response, making its reward learning curve bouncing between the range 0.003 to 0.001, making the model hard to learn more meaningful 

We examine the reward learning curves of models trained with the four types of rewards.
Reward curves in Figure \ref{fig:training_curves} during \grpo{} training show key differences across reward functions. Traditional metrics show minimal reward change, only 0.05 for \rouge{}-L and 0.25 for \bertscore{}, indicating limited model improvement across all global training steps. In contrast, 3B-\grmLlama{}'s reward is strongly correlated with response length; by step 60, it already generates the maximum allowed tokens (1,024), causing reward values to plateau around 0.6. \preferenceBert{} shows a more meaningful reward progression, not strictly tied to length, suggesting it favors responses of an optimal length rather than simply longer outputs.

\section{Human Evaluation}
\label{sec:human_eval}

While \llm{}-as-a-judge evaluation often correlates with human judgments at the system level~\cite{gu2025surveyllmasajudge}, \llm{}s tend to prefer longer responses regardless of quality~\cite{zheng2023judgingllmasajudgemtbenchchatbot}.
To better assess output quality, we conducted a human evaluation of responses from seven models: Qwen2.5-72B-Instruct, Qwen2.5-3B-Instruct, 3B-\grmLlama{}, 3B-\preferenceBert{}, 3B-RougeL, 3B-\bertscore{}, and 3B-\sft{}.
Human preferences largely align with \lmJudge{} rankings, except for 3B-\grmLlama{}, whose overly verbose answers lower its human scores.
Among the 3B models, 3B-\preferenceBert{} is the top performer.
Additionally, we observe that \sft{} models often produce shallow outputs, whereas \grpo{} models, trained with strong verifiable reward signals, better leverages the model's internal capabilities to produce higher-quality responses.

\subsection{Human Evaluation Setup}
We randomly sample 150 test prompts (50 from each dataset's test set), then collect responses from Qwen2.5-72B-Instruct, Qwen2.5-3B-Instruct, 3B-\grmLlama{}, 3B-\preferenceBert{}, 3B-RougeL, 3B-\bertscore{}, and 3B-\sft{}.
% We sample 50 prompts from the test set of each dataset, with a total of 150 prompts across three datasets, where each prompt has seven responses from our selected models.
%
We use an annotation tool (\S.Figure~\ref{fig:annotation_app}), where for each response, the annotator needs to give a Likert score between 1-5 using the same evaluation criteria as the criteria given to \llm{}s in \S Table~\ref{tab:prompt_templates_pointwise_eval}.
We have four author annotators annotating a total of 150 examples.
For each prompt, the annotator also needs to give rankings of the responses of the seven models. 
All the model names are hidden for a fair comparison.

\subsection{Results and Qualitative Analysis} \label{subsec:qual_analysis}
% Table~\ref{tab: humanEval} shows that Qwen2.5-72B-Instruct has the highest average human ratings and success rates-- 3.61 and 57\%, followed by \preferenceBert{}--3.36 and 51\%. 
% %
% On the other hand, 3B-\bertscore{} and 3B-\rouge{}-L are slightly better than the 3B base model by $\approx3\%$ success rate where 3B-\preferenceBert{} has $\approx10\%$. 3B-\sft{} has the lowest rating and 3B\grmLlama{} has the second lowest rating and win rates.
% %
% Besides 3B-\grmLlama{}, our human evaluation results are consistent with results of \lmJudge{} in terms of model ranking.
% %
% For the next section, we run qualitative analysis and case studies that reveal our three findings.
%

\begin{table*}[ht]
  \tiny
  \centering
  \resizebox{\textwidth}{!}{%
  \begin{tabular}{%
      >{\raggedright\arraybackslash}p{2.2cm} % Model column
      rrrr|rrrr|rrrr                         % means | success | BT scores
    }
    \toprule
    & \multicolumn{4}{c|}{\bfseries Mean Likert Scores} 
    & \multicolumn{4}{c|}{\bfseries Success Rates with Score $\geq 4$ (\%)}
    & \multicolumn{4}{c}{\bfseries Bradley–Terry Win Rate (\%)} \\
    \cmidrule(lr){2-5}\cmidrule(lr){6-9}\cmidrule(lr){10-13}
    Model 
      & \el{}      & \longform{}  & \alpaca{}  & Overall 
      & \el{}      & \longform{}  & \alpaca{}  & Overall 
      & \el{}      & \longform{}  & \alpaca{}  & Overall  \\
    \midrule
    % \midrule
    \multicolumn{13}{c}{Base \llm{}} \\
    \hline
    Qwen2.5-72B-Instruct   & 3.85 & 3.9 & 3.4 & 3.61 & 70.0 & 65.0 & 47.5 & 57.3 & 16.67 & 21.54 & 17.62 & 17.62 \\
Qwen2.5-3B-Instruct    & 3.31 & 3.3 & 3.2 & 3.21 & 40.0 & 55.0 & 30.0 & 37.80 & 15.24 & 14.06 & 12.60 & 12.62 \\
\midrule
    \multicolumn{13}{c}{\rl{} Finetuned Policy Models (\grpo{})} \\
    \hline
    3B-\grmLlama{}         & 2.30 & 2.5 & 2.75 & 2.55 & 15.0 & 30.0 & 47.5 & 31.7 & 12.86 & 8.16 & 10.76 & 14.76 \\
3B-\preferenceBert{}   & 3.55 & 3.6 & 3.5 & \cellcolor{cyan!20}3.36 & 60.0 & 55.0 & 55.0 & \cellcolor{cyan!20}51.0 & 21.19 & 19.72 & 21.38 & \cellcolor{cyan!20}17.38 \\
3B-\bertscore{}        & 2.95 & 3.3 & 3.3 & 3.23 & 40.0 & 45.0 & 42.5 & 41.46 & 15.95 & 12.02 & 17.62 & 16.43 \\
3B-\rouge{}-L          & 3.40 & 2.9 & 3.3 & 3.31 & 53.0 & 43.5 & 27.5 & 41.66 & 15.24 & 19.04 & 16.43 & 17.14 \\
\midrule
    \multicolumn{13}{c}{Supervised Finetuning (\sft{})} \\
    \hline
3B-sft                 & 2.0 & 2.8 & 1.4 & 1.93 & 10.0 & 25.0 & 10.0 & 13.41 & 0.03 & 5.44 & 3.98 & 4.05 \\

    \bottomrule
  \end{tabular}
  }
  \caption{Human evaluation shows a different preference than automatic evaluation, where model trained with \grmLlama{} is less preferred by exerts but \preferenceBert{} is more preferred than all other policy models.}
  \label{tab: humanEval}
\end{table*}

\begin{table}[ht]
  \centering
  \tiny
\resizebox{\columnwidth}{!}{%
  \begin{tabular}{lrrr}
    \toprule
    Model 
      & \makecell{Markdown\\(\%)} 
      & \makecell{Repetition\\Rate (\%)} 
      & \makecell{Response\\Length} \\
    \midrule
    Qwen2.5-72B-Instruct     & 47.48 & 6.25 & 220\\
    Qwen2.5-3B-Instruct      & 28.89 & 4.69 & 194\\
    3B-\grmLlama{}           & 66.80 & 4.18 & \textbf{710} \\
    3B-\preferenceBert{}     & 81.31 & 4.34 & 258 \\
    3B-\bertscore{}          & 24.00 & 4.55 & 180\\
    3B-\rouge{}-L            & 21.92 & 8.59 & 182\\
    3B-\sft{}                & 15.96 & 8.29 & 146 \\
    \bottomrule
  \end{tabular}
}
\caption{Average words per response for each group by model. 3B-\grmLlama{} generates way more words per response than all other models, where human experts consider as overly verbose and contain unnecessary information. Repetition rate is the percentage of bigrams that are repeated. Markdown is a regular expression that checks whether a response follows a particular structure and returns a boolean (\S Table~\ref{tab:markdown-expression-check}).}
\label{tab:avg-words-per-response}
\end{table}

Table~\ref{tab: humanEval} shows that Qwen2.5-72B-Instruct achieves the highest average human Likert rating (3.61) and success rate (57\%), followed by 3B-\preferenceBert{} with a rating of 3.36 and a 51\% success rate.
In contrast, 3B-\bertscore{} and 3B-\rouge{}-L perform only slightly better than the 3B base model, each improving success rates by approximately 3\%. 3B-\preferenceBert{}, however, achieves a much larger gain of around 10\%.
3B-\sft{} receives the lowest human rating, while 3B-\grmLlama{} ranks second to last in both Likert score and win rate.
Except for 3B-\grmLlama{}, the human evaluation results align closely with the rankings produced by \lmJudge{}, supporting the reliability of our automatic evaluation.
We elaborate on these human evaluation findings below.
%
% In the next section, we present a qualitative analysis and case studies that highlight three key findings behind these performance differences.

% \paragraph{While \llm{} judges favor 3B-\grmLlama{}, annotators strongly prefer 3B-\preferenceBert{}.}
\paragraph{Annotators strongly prefer 3B-\preferenceBert{}.}
In general, we observe a consistent ranking between human expert judgments and the rankings produced by LLM-as-a-judge. However, 3B-\grmLlama{} is a notable outlier.
While it has the highest scores from \llm{} judges across all \grpo{}-trained models (Table~\ref{tab:combined_full_performance}), annotators consistently rank it low (Table~\ref{tab: humanEval}). As shown in Table~\ref{tab:avg-words-per-response}, 3B-\grmLlama{} is excessively verbose, averaging 710 words per response—almost 2.75 times as many as 3B-\preferenceBert{}, which averages 258.
%
% In addition, we further discuss how \grmLlama{} correlates response length with reward during training, which could cause the model to systematically favor long outputs.
% The long response length is also consistent with the reward training curves in \S Figure~\ref{fig:training_curves}, that \grmLlama{} is purely rewards based on the length of the responses but not the actual quality, which can trick \lmJudge{}, but humans dislike the answers because of inconciseness.
The tendency for longer responses is consistent with the reward training curves in Figure~\ref{fig:training_curves}: 3B-\grmLlama{} receives higher rewards primarily for increased response length, rather than for high answer quality. While this reward hacking can lead \lmJudge{} to assign higher scores, human evaluators often dislike these responses due to their verbosity and lack of conciseness and completeness.
Additionally, we observe that 3B-\grmLlama{} requires 10 more hours of training than other 3B policy models, and 1.5B-\grmLlama{} takes 7 more hours than other 1.5B models. This increased training time is a result of length-based rewards, which guide the model to produce longer responses—thereby increasing the time needed for each \grpo{} inference and overall training.

\paragraph{What distinguishes 3B-\preferenceBert{} from the base model?}
We further analyze 26 of the examples where 3B-\preferenceBert{} is preferred over the base model. The improvements fall into two main categories: \textbf{instruction following} and \textbf{tone and fluency}. On prompts with explicit constraints (e.g., “explain in 2 sentences”), the base model often fails to comply, producing responses that are either too long or overly brief. In contrast, 3B-\preferenceBert{} reliably adheres to such constraints. Additionally, its writing is more polished and human-like. While the base model tends to sound mechanical—producing fragmented sentences reminiscent of stitched-together search results—3B-\preferenceBert{} generates fluent, cohesive answers. See Appendix Figure~\ref{showcases:instruct_following} and Figure~\ref{showcases:tone_and_fluency} for qualitative examples.

\paragraph{3B-\preferenceBert{} and 3B-\grmLlama{} frequently use structured outputs.}
In Table~\ref{tab:avg-words-per-response}, we observe that 3B-\preferenceBert{} and 3B-\grmLlama{} tend to use markdown formatting more frequently than other models (81.31\% and 66.80\%, respectively).
As a result, annotators often note that their outputs have better \textbf{readability}. See qualitative examples in Appendix (Figure~\ref{showcases:readability}).

\paragraph{3B-\rouge{}-L and 3B-\bertscore{} tend to be generic and sometimes repetitive.}
These models often respond to prompts such as \textit{“Categorize the AI technologies mentioned below: Machine Learning, Natural Language Processing, Robotics”} with generic definitions (e.g., \textit{“Machine Learning is a subset of artificial intelligence that involves training algorithms...”}) rather than actually categorizing or differentiating between the terms.
Additionally, Table~\ref{tab:avg-words-per-response} shows that 3B-\rouge{}-L can be highly repetitive.
In \S\ref{app:reward_curves}, we find that \rouge{}-L and \bertscore{} show little reward variance across the training curve, which could indicate that the training signal is too weak, potentially causing the trained models to output vague and surface-level outputs.
In contrast, 3B-\preferenceBert{} provides clearer categorizations and contextual explanations for each term, demonstrating stronger \textbf{content logic}.
See detailed qualitative analysis in Appendix Figure~\ref{showcases:content_logic} and Figure~\ref{showcases:countercase_sft}.
% \paragraph{3B-\sft{} responses are often vague, surface-level, or evasive, which leads to lower human preference ratings.}
\paragraph{3B-\sft{} responses are often vague and overly simplified.}
In annotated examples, 3B-\sft{} responses explicitly avoid answering the question—sometimes stating “I don't know” or offering no meaningful explanation. For instance, in response to the prompt \textit{“Why is the Big Bang seen as a singular event?”}, the model deflects the question without addressing the core scientific reasoning.
Additionally, on \longform{} prompts—especially those derived from Alpaca-style or open-ended datasets—3B-\sft{} tends to produce overly simplified, shallow explanations. These responses often lack both technical depth and structural clarity, which diminishes their informativeness and readability.
This trend is also reflected in Table~\ref{tab:avg-words-per-response}, where 3B-\sft{} produces the shortest responses on average.
We attribute this issue in part to the nature of the training data from sources such as \el{}, which contains casual, informal responses—many of which may be low-quality or factually incorrect. This results in a model that mimics the tone and content of noisy or imprecise reference answers.
While \grpo{}-trained models demonstrate better performance over \sft{} in open-ended long-form generation in our experiments, we do not dismiss \sft{} as an ineffective approach. When high-quality, human-annotated datasets are available, \sft{} remains a valuable strategy—particularly in domains like code generation~\cite{zhou2023limaalignment}, where reference outputs are well-defined and reliable.

\section{Conclusion}
\label{sec:con}
\rlvr{} especially \grpo{} has been a success for its ability to fully leverage \llm{}s' abilities to self-improve without massive amount of labeled data on many rule-based evaluation tasks.
However, extending \grpo{} study on long-form and open-ended generation has been underexplored for the challenges of evaluating long-form responses.
We propose using a small fine-tuned language model (\preferenceBert{}) to evaluates long-form responses with different dimensions and semantic quality evaluation as reward signals for long-form generations of \grpo{} and show that model trained with \preferenceBert{} generates responses with overall better quality than models trained with traditional metrics such as \rouge{} and \bertscore{} or a generalized preference reward model, even close to the quality of larger models of the same backbone.
Our work shows the potential of applying more efficient and robust verifiable reward design into the \grpo{} pipeline for models to self-improve its long-form and open-ended generations beyond using traditional reward metrics. 
Future work can expand upon current work on more diverse open-ended generation tasks such as training more efficient and stronger verifiable reward models and apply them on creative writings, creative research and design, or open-ended math problems.

\section{Ethics}
Our annotation does not involve in collecting annotators' private information and does not involve in extensive tool usage.
Thus, our annotation is exempted by the Institutional Review Board (\abr{irb}) annotation protocol.

\section{Limitations}
\label{sec:limitations}
We are the first to demonstrate that fine-tuned evaluation language models can effectively leverage the capabilities of \llm{}s for evaluating and improving long-form, open-ended generations.
However, this work has not fully explored the potential of \grpo{} and reward design for enabling self-improvement in \llm{}s on such complex tasks.
A key limitation of our study is that we did not train or use a larger and more powerful language model (e.g., 7B-scale) to serve as a verifiable reward provider, primarily due to computational constraints.
Larger evaluators, while potentially offering more reliable and semantically accurate rewards, significantly increase GPU memory usage and slow down training.
We hypothesize that incorporating a stronger evaluator to provide high-quality, verifiable rewards could unlock the full potential of \grpo{} for aligning \llm{}s on open-ended tasks.

% Preprocessing

% stage 2
\bibliography{bib/2025_arr_GTD}

\clearpage
\appendix

% \section{Appendix}
% \label{sec:appendix}
% % \input{figures/var}
\section{Technical Details}
\label{app:tech}
% In this section, we provide supplemental technical details for \grpo{} mentioned in Section~\ref{sec:reward_model} to further illustrating the regularization terms used \grpo{} and the details of Bradley and Terry Loss used by \grm{}.
In this section, we provide additional technical details for \grpo{} mentioned in Section~\ref{sec:reward_model}, further illustrating the regularization terms used in \grpo{} and the specifics of the Bradley-Terry loss employed by \grm{}.
\subsection{\grpo{} Regularization Objective}
\begin{multline}
\label{eq:grpo_objective_detailed}
\mathcal{J}_{\text{GRPO}}(\phi) = \mathbb{E}_{x, \{y_i\}} \Biggl[ \frac{1}{G} \sum_{i=1}^{G} \min\Biggl( \rho_i(\phi) A(x,y_i), \\
\text{clip}\left(\rho_i(\phi), 1-\epsilon, 1+\epsilon\right) A(x,y_i) \Biggr) \Biggr] \\ - \beta \, \mathbb{E}_{x \sim \mathcal{D}}[\text{KL}(\pi_{\phi}(\cdot|x) || \pi_{\text{ref}}(\cdot|x))],
\end{multline}

where $\rho_i(\phi) = \frac{\pi_\phi(y_i \mid x)}{\pi_{\phi_{\text{old}}}(y_i \mid x)}$ is the probability ratio for $y_i$, $\epsilon$ is the clipping hyperparameter and $\beta$ is the KL penalty coefficient.
% \yapei{can move formula to appendix and just describe in words and symbols; the important idea is the group normalization part}

\subsection{\grm{} Bradley and Terry Loss}
The \grmLlama{} is trained to minimize the Bradley-Terry Loss:

\begin{equation}
\label{eq:grmBtLoss}
\resizebox{\linewidth}{!}{$
L_{\text{reward}}(\theta) = - \mathbb{E}_{(x, y_c, y_r)} \left[ \log \sigma \left( r_\theta(x, y_c) - r_\theta(x, y_r) \right) \right]
$}
\end{equation}
where $r_\theta(x, y)$ denotes the reward score predicted by the model and $\sigma(\cdot)$ is the sigmoid function.
%
% The \grm{} incorporates a text-generation regularization term that preserves the language modeling capabilities of the hidden states:
% \begin{equation}
% L_{\text{total}} = (1 - \alpha) L_{\text{reward}} + \alpha L_{\text{reg}},
% \end{equation}
% where $L_{\text{reg}}$ is a text-generation loss such as supervised fine-tuning (SFT) loss, and $\alpha$ is a balancing coefficient.
% Once trained, the reward model $r_\theta(x, y)$ is used to evaluate and rank responses generated by a language model, either for selection in Best-of-$n$ (BoN) decoding or as the optimization objective in reinforcement learning (e.g., PPO).
%
Generally, the \grm{} $r_\theta(x, y)$ is used in \abr{rlhf} training, either for selection in Best-of-$n$ (BoN) decoding or as the optimization objective in reinforcement learning (e.g., PPO~\cite{schulman2017proximalpolicyoptimizationalgorithms}).
%
% In our scenario, we neither use \grmLlama{} to rank models nor use it as a generalized PPO reward model, we use it as a \grm{} as reward signals for \grpo{}.
% %
% Specifically, since the output range of the model is real-numbers, to condense the reward signals, we apply a sigmoid function to normalize the rewards to be between 0 and 1.
% %
% We use \grmLlama{} because it is one of the best models under 7B on rewardBench~\cite{lambert2024rewardbenchevaluatingrewardmodels}, where a large model can consume significant more amount of GPU memory during training, not only slowing down the process, but may also requires additional computation memory that is not available to everyone.
We use \grmLlama{} as a \grm{} to provide reward signals for \grpo{}, rather than for ranking models or as a \abr{ppo} reward model. 
We use a sigmoid function to normalize its real-valued outputs to the [0, 1].\footnote{We choose \grmLlama{} for its best performance as the smallest model on RewardBench~\cite{lambert2024rewardbenchevaluatingrewardmodels}, which offers a good trade-off between quality and efficiency without the heavy GPU demands of larger models.}

\begin{table*}[ht]
\centering
\scriptsize
\begin{tabular}{m{0.11\textwidth} m{0.08\textwidth} m{0.08\textwidth} m{0.29\textwidth} m{0.29\textwidth}}
\toprule
\textbf{Dataset} & \textbf{\# Train} & \textbf{\# Test} & \textbf{Example Input} & \textbf{Example Reference Response} \\
\midrule
\abr{\el{}} & 10,444 & 1,056 & Could we theoretically create an infinite echo? & The perfect conditions would be a wall of atoms that will not move at all when bumped. Considering the fact that heat is defined by the movement of atoms... \\
\midrule
\longform{} & 8,648 & 956 & Explain how Venezuela raised its minimum wage. & Venezuela raised its minimum wage to 1 million bolivars per month on Monday, the third increase this year that puts the figure at just \$1.61 at the black market exchange rate. President Nicolas Maduro... \\
\midrule
Alpaca & 10,444 & 1,334 & Develop a customer service strategy to improve customer experience. & Here is a customer service strategy that can help in improving the customer experience: 1. Identify your customers' needs... \\
\bottomrule
\end{tabular}
\caption{Overview of the datasets used in our experiments. All datasets contain long-form, open-ended questions spanning diverse domains (e.g., science, instruction following), with responses averaging 185 words.}
\label{tab:dataset}
\end{table*}

\section{Dataset Details} \label{app:dataset_details}
% We provide the details over the datasets we used in our work in Table~\ref{tab:dataset}, including the size of both training and testing dataset, along with the example inputs and reference response involved.
Table~\ref{tab:dataset} presents details of the datasets used in our work, including the sizes of the training and testing sets, as well as example inputs and reference responses.
% \paragraph{Explain Like I'm 5 (\el{})} is a collection of questions and answers from Reddit's r/explainlikeimfive community~\cite{fan2019eli5longformquestion}.\footnote{\url{https://www.reddit.com/r/explainlikeimfive/}} 
% %
% It contains 270K threads where people ask open-ended questions, and others respond with simple, easy-to-understand explanations, as if explaining to a five-year-old across areas e.g. like chemistry, psychology, biology, and earth science. 
% % 
% Its goal is like teaching a language model to explain things in a way that's easy for everyone to understand .
% %
% We sample 10,444 questions as the training set and 1,056 as the test set. 
\paragraph{Explain Like I'm 5 (\el{})} is a dataset derived from Reddit’s r/explainlikeimfive community~\cite{fan2019eli5longformquestion}.\footnote{\url{https://www.reddit.com/r/explainlikeimfive/}} It contains 270K threads where users ask open-ended questions and receive simple, easy-to-understand explanations—framed as if explaining to a five-year-old. The topics span a wide range of domains, including chemistry, psychology, biology, and earth science. The dataset is intended to help models learn to explain complex topics in accessible ways. We sample 10,444 questions for training and 1,056 for testing.

% \paragraph{\alpaca{}} is a collection of 52K instruction-response pairs generated by OpenAI’s text-davinci-003 to fine-tune LLaMA 7B~\cite{alpaca2023}.\footnote{\url{openai.com}}
% %
% It diverse prompts and corresponding long responses in the style of the Self-Instruct~\cite{selfinstruct}. 
% %
% We adopt the cleaned version of \alpaca{}~\cite{alpaca} that removes examples with original hallucinating answers, empty responses, instruction to generate images.
% %
% We remove examples that have response length fewer than 50 words and sample 10,444 examples as train set and 1,334 examples as the test set.
\paragraph{\alpaca{}} is a collection of 52K instruction-response pairs generated by OpenAI’s text-davinci-003 to fine-tune the LLaMA 7B model~\cite{alpaca2023}.\footnote{\url{https://openai.com}}
It features diverse prompts and long-form responses in the style of Self-Instruct~\cite{selfinstruct}.
We use a cleaned version of \alpaca{}~\cite{alpaca} that removes instances with hallucinated answers, empty responses, or instructions to generate images.
Additionally, we filter out examples with responses shorter than 50 words, resulting in a final set of 10,444 training and 1,334 test examples.

% \paragraph{\longform{}} is created by leveraging English corpus examples with reverse instructions~\cite{koksal2023longform}.
% %
% It contains diverse set of human-written documents from e.g. Wikipedia~\cite{wikipedia}, C4~\cite{c4}, Stack Exchange~\cite{stackexchange2025}, Big Bench~\cite{bigbench} and the instructions are generated via \llm{}s with task examples spanning from question answering, email writing, story/poem generation, and text summarization.
% %
% We remove the examples that requires coding from our examples since it is considered out of scope and we sample 8,648 questions as the training set and 956 as the test set.
%

\paragraph{\longform{}} is constructed by applying reverse instruction generation to an English corpus, following the approach in~\cite{koksal2023longform}.
It includes a diverse set of human-written documents sourced from Wikipedia~\cite{wikipedia}, C4~\cite{c4}, Stack Exchange~\cite{stackexchange2025}, and BigBench~\cite{bigbench}. Instructions are generated by \llm{}s, covering a wide range of tasks such as question answering, email writing, story or poem generation, and text summarization.
We exclude examples requiring code generation, as they fall outside our intended scope. The final dataset contains 8,648 training examples and 956 test examples.

% \section{Reward Learning Curves} \label{app:reward_curves}
% % \textcolor{red}{Discuss the reward curves}
% % due to \rouge{}-l' lack of semantic evaluation property, it cannot give meaningful reward feedback signal to the model, and the model-generated responses are quite different from the reference response, making its reward learning curve bouncing between the range 0.003 to 0.001, making the model hard to learn more meaningful 

% We examine the reward learning curves of models trained with the four types of rewards.
% %
% Reward curves in Figure \ref{fig:training_curves} during \grpo{} training show key differences across reward functions. Traditional metrics show minimal reward change, only 0.05 for \rouge{}-L and 0.25 for \bertscore{}, indicating limited model improvement across all global training steps. In contrast, 3B-\grmLlama{}'s reward is strongly correlated with response length; by step 60, it already generates the maximum allowed tokens (1,024), causing reward values to plateau around 0.6. \preferenceBert{} shows a more meaningful reward progression, not strictly tied to length, suggesting it favors responses of an optimal length rather than simply longer outputs.

\section{Prompt Template}
\label{app:eval_prompt_template}
% We show prompt template used for training in Table~\ref{tab:training_prompt_temps}, the prompt template for both point-wise evaluation in Table~\ref{tab:prompt_templates_pointwise_eval} and pairwise-preference evaluation in Table~\ref{tab:prompt_templates_pairwise_eval}.

We show the prompt template used for training in Table~\ref{tab:training_prompt_temps}, the template for point-wise evaluation in Table~\ref{tab:prompt_templates_pointwise_eval}, and the template for pairwise preference evaluation in Table~\ref{tab:prompt_templates_pairwise_eval}.

\begin{table}[h]
\small
\setlength{\tabcolsep}{4pt}
\centering
\resizebox{1.0\columnwidth}{!}{
\begin{tabular}{c p{12cm}}
 \toprule
& \multicolumn{1}{c}{\bf Training Prompt Template} \\
\midrule
\noalign{\vskip 1mm}
& {\ttfamily
The user asks a question, and the Assistant answers it. The assistant provides the user with the answer that strictly follows the following guidelines. The answer should be enclosed within <answer> </answer> tags, respectively, i.e., <answer> ANSWER HERE </answer>. Your answer should follow these rubric criteria: \par
Rubric: \par
Factual Accuracy: The answer must be factually correct and does not contradict the reference answer. \par
Relevance and Completeness: The answer should directly address the specific question, covering all essential aspects. \par
Clarity and Organization: The answer should be well-structured, coherent, and easy to follow. \par
Conciseness: The answer should avoid unnecessary repetition and be as clear and succinct as possible. \par
Completeness: The answer is complete and not repetitive. \par
Response Format rules:\par
- Always start your response with <answer> tag and end with </answer>.\par
- Do not include any text or commentary before the opening <answer> tag and after the closing </answer> tag.\par
For example, your response should follow this format:\par
<answer>\par
[Your final detailed answer goes here]\par
</answer>\par
Question: \{question\}
} \\
\noalign{\vskip 1mm}
\bottomrule
\end{tabular}
}
\caption{Training prompt template for \texttt{LLM}s to generate responses.}
\label{tab:training_prompt_temps}
\vspace{-10pt}
\end{table}

\section{Markdown Expression Check} \label{app:markdown_expression}
We provide Table~\ref{tab:markdown-expression-check} to exhibit the markdowns, which is a regular expression that
checks whether a response follows a particular structure
% and returns a boolean, as we mentioned in the Caption of Table~\ref{tab:avg-words-per-response}.
We provide Table~\ref{tab:markdown-expression-check} to present the markdown patterns—regular expressions used to check whether a response follows a specific structure and return a boolean value, as described in the caption of Table~\ref{tab:avg-words-per-response}.

\begin{table}[ht]
  \centering
  \footnotesize
  \caption{Markdown Expression Check}
  \label{tab:markdown-expression-check}
  \begin{tabular}{@{}ll@{}}
    \toprule
    \textbf{Pattern} & \textbf{Description} \\
    \midrule
    \verb|^#{1,6}\s|       & ATX-style headings (e.g. \texttt{\# Heading}) \\
    \verb|^[-*+]\s|        & Unordered list items (e.g. \texttt{- item}) \\
    \verb|^\d+\.\s|        & Ordered list items (e.g. \texttt{1. item}) \\
    \verb|^>\s|            & Blockquotes (e.g. \texttt{> quote}) \\
    \verb|```[\s\S]+?```|  & Fenced code blocks \\
    \verb|`[^`\n]+?`|      & Inline code spans \\
    \verb!\\|.+\\|!        & Tables using pipe separators \\
    \bottomrule
  \end{tabular}
  \vspace{-10pt}
\end{table}

\section{Annotation Tool}
\label{app:annotation_tool}
\begin{figure*}[htbp] % position: here, top, bottom, page
    \centering
    \includegraphics[width=0.7\textwidth]{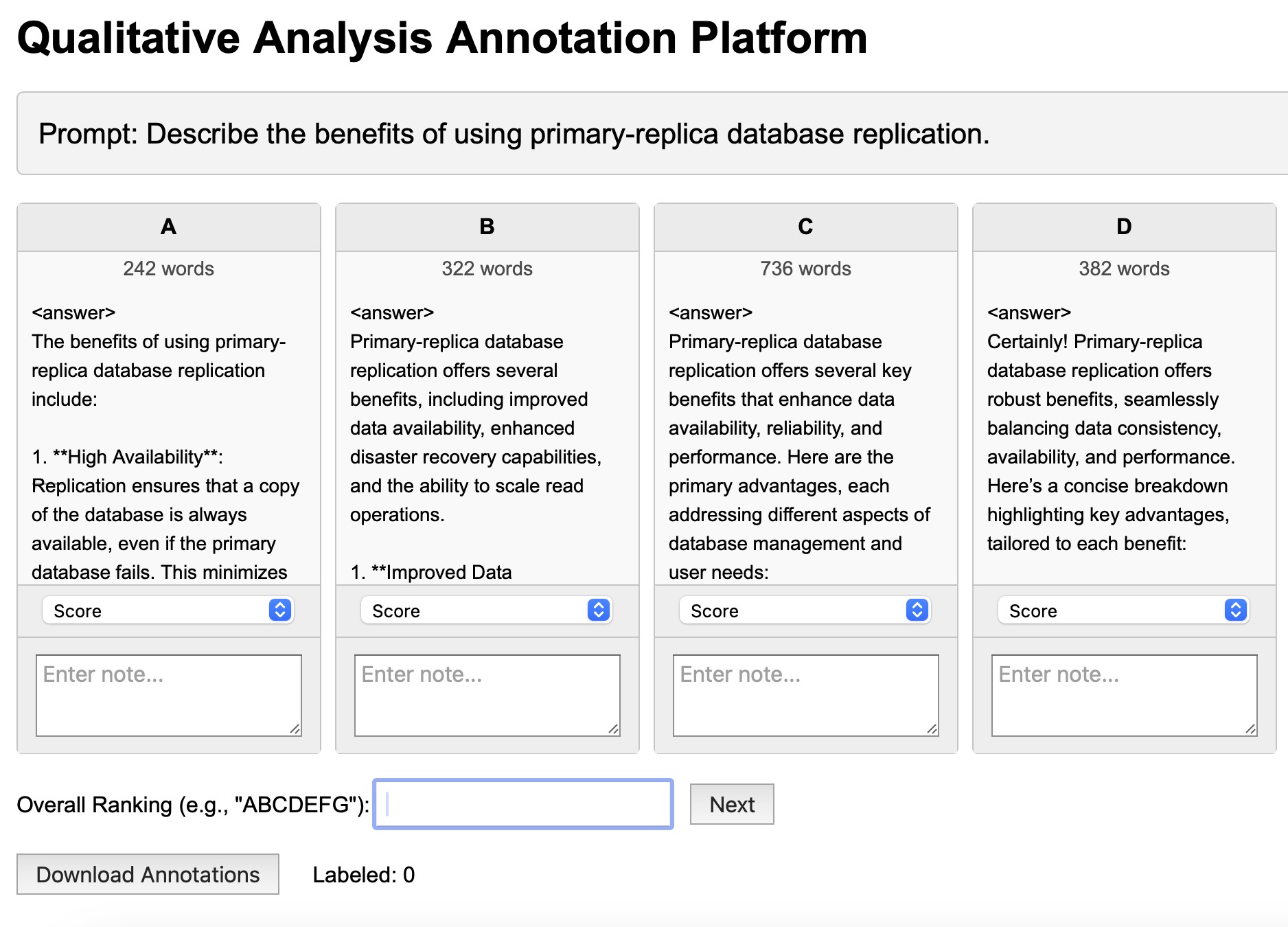}
    \caption{Our annotation tool for response quality annotation. Annotators will be displayed with the question prompt, the answers for the seven models, where they need to slide due to limited screen width. Annotators can then put their Likert scores (1-5) and comments or notes for each response, and then finally rank the responses based on their preferences and ratings.}
    \label{fig:annotation_app}
    \vspace{-10pt}
\end{figure*}

We present the annotation tool for response quality annotation in Figure~\ref{fig:annotation_app}. Annotators will be displayed with the question prompt, the answers for the seven models, where they need to slide due to limited screen width. Annotators can then put their Likert scores (1-5) and comments or notes for each response, and then finally rank the responses based on their preferences and ratings.

\section{Showcases for Qualitative Analysis} \label{app:qual_examples}
% Given the criteria in Section \ref{subsec:qual_analysis}, we provide multiple examples showcasing the critiat we use to evualate the quality of the free-form long-context generation results. Our showcases for qualitative analysis are provided as follows.
We present qualitative examples to accompany our analysis in Section~\ref{subsec:qual_analysis}. See Figure~\ref{showcases:readability} (Readability), Figure~\ref{showcases:content_logic} (Content Logic), Figure~\ref{showcases:instruct_following} (Instruction Following), Figure~\ref{showcases:tone_and_fluency} (Tone and Fluency), and Figure~\ref{showcases:countercase_sft} (Failure cases of SFT).

\begin{table}[h]
\small
    \setlength{\tabcolsep}{4pt}
    \centering
    \resizebox{1.0\columnwidth}{!}{
    \begin{tabular}{c p{9cm}}
     \toprule
    & \multicolumn{1}{c}{\bf Point-wise Evaluation Template} \\
    \midrule
     \noalign{\vskip 1mm}
     & \texttt{You will be given a user question, a reference answer, and a system answer. Your task is to provide an overall rating scoring how well the system answer addresses the user question against the reference answer. Give your answer as an integer on a scale of 1 to 5, where 1 means that the system answer is not informative, and 5 means that the answer addresses the question according to the criteria below.}\\
     \noalign{\vskip 2mm}
     & \texttt{Rubric:} \\
     \noalign{\vskip 2mm}
     & \texttt{Factual Accuracy: The answer must be factually correct and does not contradict the reference answer.} \\
    \noalign{\vskip 2mm}
     & \texttt{Relevance and Completeness: The answer should directly address the specific question, covering all essential aspects.} \\
    \noalign{\vskip 2mm}
     & \texttt{Clarity and Organization: The answer should be well-structured, coherent, and easy to follow.} \\
     \noalign{\vskip 2mm}
     & \texttt{Conciseness: The answer should avoid unnecessary repetition and be as clear and succinct as possible.} \\
     \noalign{\vskip 2mm}
     & \texttt{Completeness: The answer is complete and not repetitive.} \\
     \noalign{\vskip 2mm}
     & \texttt{Please base your overall rating on how well the system answer performs in these areas.} \\
     \noalign{\vskip 2mm}
     & \texttt{Question: \{question\}} \\
     \noalign{\vskip 2mm}
     & \texttt{Reference Answer: \{reference\_answer\}} \\
     \noalign{\vskip 2mm}
     & \texttt{System Answer: \{answer\}} \\
     \noalign{\vskip 2mm}
     & \texttt{Please be as strict and as critical and harsh as possible. } \\
     \noalign{\vskip 2mm}
     & \texttt{Provide your feedback as follows:} \\
     \noalign{\vskip 2mm}
     & \texttt{Feedback:::} \\
     \noalign{\vskip 2mm}
     & \texttt{Final rating: (your rating, as an integer between 1 and 5)} \\
     \noalign{\vskip 1mm}
    \bottomrule
    \end{tabular}
    }
    \caption{Prompt template for point-wise evaluation.} 
    \label{tab:prompt_templates_pointwise_eval}
\end{table}

\begin{table}[h]
\small
\setlength{\tabcolsep}{4pt}
\centering
\resizebox{1.0\columnwidth}{!}{
\begin{tabular}{c p{12cm}}
 \toprule
& \multicolumn{1}{c}{\bf Pairwise Preference Evaluation Template} \\
\midrule
\noalign{\vskip 1mm}
& {\ttfamily
You are a fair judge assistant tasked with providing clear, objective feedback based on specific criteria, ensuring each assessment reflects the absolute standards set for performance. \par
Your task is to provide your preferred response as either A or B. Please strictly follow the output format as: \par
Feedback: Reason why you choose this answer \par
[RESULT] A or B</s> \par
Rubric: \par
Factual Accuracy: The answer must be factually correct and does not contradict the reference answer. \par
Relevance and Completeness: The answer should directly address the specific question, covering all essential aspects. \par
Clarity and Organization: The answer should be well-structured, coherent, and easy to follow. \par
Conciseness: The answer should avoid unnecessary repetition and be as clear and succinct as possible. \par
Completeness: The answer is complete and not repetitive. \par
Write a detailed feedback that assesses the quality of two responses strictly based on the given score rubric, not evaluating in general. \par
After writing a feedback, choose a better response between Response A and Response B. You should refer to the score rubric. \par
Question: \{question\} \par
Reference Answer: \{reference\_answer\} \par
Answer A: \{answer\_A\} \par
Answer B: \{answer\_B\} \par
Please be as strict and as critical and harsh as possible. \par
Provide your feedback as follows: \par
Feedback::: \par
Final rating: (your rating, as an integer between 1 and 5) \par
} \\
\noalign{\vskip 1mm}
\bottomrule
\end{tabular}
}
\caption{Prompt template for pairwise evaluation.} 
\label{tab:prompt_templates_pairwise_eval}
\vspace{-10pt}
\end{table}
$$$$

% \begin{figure}[htbp]
%     \centering
%     \includegraphics[width=0.46\textwidth]{figures/qual_exorcist.jpg}
%     \caption{Example of reward model misalignment: The prompt requests a 66-word response. While most completions respect the word constraint, one model generates a 688-word answer that is overly verbose. Such responses are often rewarded by large general reward models (\grm{}), despite violating task constraints.}
%     \label{fig:qual_exorcist}
% \end{figure}

% \begin{table}[ht]
%   \centering
%   \caption{Markdown Expression Check}
%   \label{tab:markdown-expression-check}
%   \begin{tabular}{@{}ll@{}}
%     \toprule
%     \textbf{Pattern} & \textbf{Description} \\
%     \midrule
%     \verb|^#{1,6}\s|       & ATX-style headings (e.g. \texttt{\# Heading}) \\
%     \verb|^[-*+]\s|       & Unordered list items (e.g. \texttt{- item}) \\
%     \verb|^\d+\.\s|       & Ordered list items (e.g. \texttt{1. item}) \\
%     \verb|^>\s|           & Blockquotes (e.g. \texttt{> quote}) \\
%     \verb|```[\s\S]+?```| & Fenced code blocks \\
%     \verb|`[^`\n]+?`|     & Inline code spans \\
%     \verb|\|.+\||         & Tables using pipe separators \\
%     \bottomrule
%   \end{tabular}
% \end{table}

\begin{figure*}
\begin{center}
\begin{tikzpicture}
\node [mybox,title=Qualitative Analysis: Readability] (box){%
    \begin{minipage}{2\columnwidth}
        \raggedright
        \bluebold{Prompt: } Describe the new functions of the Tesla Model 3, and how they improve the driving experience \\
        \bluebold{Dataset: } Instruction \\
        % \quad \\ 
        \bluebold{Models Compared: } Qwen2.5-3B-PrefBERT v.s. Qwen2.5-3B-Rouge-L \&  Qwen2.5-3B-BERTScore \\
        \quad \\ 
        \bluebold{Reference Answer (Qwen2.5-72B-Instruct) }:\\ 
        \small \texttt{The Tesla Model 3 has several advanced features and functions that significantly enhance the driving experience. Here are some key highlights:
        1. **Autopilot and Full Self-Driving (FSD) Capabilities**: ...}\\
        % \bluebold{Analysis: } \\
        \quad \\ 
        \bluebold{Qwen2.5-3B-PrefBERT} \\
        \small \texttt{Certainly! Highlighting the new functions of the Tesla Model 3 and how they elevate the driving experience showcases Tesla's commitment to innovation and user-centric design. Here’s a concise breakdown, seamlessly intertwining functionality, clarity, and depth: 1. **Autopilot Advanced Driver Assistance Systems**: ...} \\
        \normalsize \bluebold{Analysis: } This answer uses more complex language, longer sentences, and a promotional tone that may appeal to a more technical or marketing-savvy reader but reduces overall clarity and ease of comprehension.\\
        \quad \\ 
        \bluebold{Qwen2.5-3B-ROUGE-L} \\
        \small \texttt{The Tesla Model 3 has several new functions that enhance the driving experience. One of the most notable improvements is the Autopilot system, which includes features like automatic emergency braking, adaptive cruise control, and lane centering. These features help drivers maintain a safe distance from other vehicles and stay within their lane, reducing the workload and stress of driving...} \\
        \normalsize \bluebold{Analysis: } This answer presents information in dense, uniform paragraphs with minimal structural guidance, making it harder for readers to scan and retain key points. It also repeats ideas (e.g., mentioning safety features like automatic emergency braking twice) and lacks cohesion, which disrupts the logical flow.\\
        \quad \\  
        \bluebold{Qwen2.5-3B-BERTScore} \\
        \small \texttt{The Tesla Model 3 has several advanced features designed to enhance the driving experience. One of its notable improvements is the Autopilot system, which allows for semi-autonomous driving capabilities. This includes automatic steering, acceleration, and braking under certain conditions, reducing the driver's workload and enhancing safety...} \\
        \normalsize \bluebold{Analysis: } This answer presents information in long, uniform paragraphs without visual structure, making it harder for readers to quickly locate and retain key points. The dense formatting in this answer lacks clear segmentation, which forces readers to parse multiple features at once without guidance or emphasis. Additionally, the transitions between topics are minimal, and the writing style is more monotonous, reducing engagement.\\
        \quad \\  
        % \bluebold{Comments on Models' Analysis: } The models \wrong{point out many capitalization, grammatical, and pronunciation errors, along with some confusing sentences,} but the Human Developers \wrong{fail to detect these mistakes.}\\
        \quad \\ 
    \end{minipage}
};
\end{tikzpicture}
\caption{\textbf{Qualitative Analysis: Readability.} We present showcases comparing the readability of \preferenceBert{} against baseline models, accompanied by detailed analysis.}
\label{showcases:readability}
\end{center}
\end{figure*}
\begin{figure*}
\begin{center}
\begin{tikzpicture}
\node [mybox,title=Qualitative Analysis: Content Logic] (box){%
    \begin{minipage}{2\columnwidth}
        \raggedright
        \bluebold{Prompt: } Categorize the AI technologies mentioned below. Machine Learning, Natural Language Processing, Robotics \\
        \bluebold{Dataset: } Alpaca Cleaned \\
        % \quad \\ 
        \bluebold{Models Compared: } Qwen2.5-3B-PrefBERT v.s. Qwen2.5-3B-Rouge-L \&  Qwen2.5-3B-BERTScore \\
        \quad \\ 
        \bluebold{Reference Answer (Qwen2.5-72B-Instruct) }:\\ 
        \small \texttt{Machine Learning is a subset of artificial intelligence that involves algorithms and statistical models that enable computer systems to improve their performance on a specific task over time with experience. Natural Language Processing (NLP) is a field of AI that focuses on the interaction between humans and computers using natural language, aiming to read, decipher, understand, and make sense of human language in a valuable and meaningful way. Robotics is a branch of AI that deals with the design, construction, operation, and use of robots, as well as computer systems for their control, sensory feedback, and information processing.}\\
        % \bluebold{Analysis: } \\
        \quad \\ 
        \bluebold{Qwen2.5-3B-PrefBERT} \\
        \small \texttt{Certainly! Categorizing the AI technologies mentioned—Machine Learning (ML), Natural Language Processing (NLP), and Robotics—brings insightful clarity to their roles and intersections within broader AI landscapes. Here’s a nuanced breakdown: 1. **Machine Learning (ML)**:... } \\
        \normalsize \bluebold{Analysis: } This answer not only defines Machine Learning, Natural Language Processing, and Robotics clearly, but also organizes the explanation into a structured, interconnected framework. It highlights how each domain functions independently while also emphasizing their synergies, particularly how Machine Learning underpins both NLP and Robotics, and how NLP enables human-machine interaction in robotic systems. This layered, relational presentation offers a more comprehensive and coherent understanding of the AI landscape\\
        \quad \\ 
        \bluebold{Qwen2.5-3B-ROUGE-L} \\
        \small \texttt{Machine Learning is a subset of artificial intelligence that involves training algorithms to make predictions or decisions without being explicitly programmed. Natural Language Processing is a field within AI that focuses on the interaction between computers and humans using natural language. Robotics is the design, construction, and operation of robots, often involving AI to enable them to perform tasks autonomously or with human assistance.} \\
        \normalsize \bluebold{Analysis: } This answer provides basic, isolated definitions of Machine Learning, Natural Language Processing, and Robotics without addressing the instruction’s likely intent: to offer an integrated and analytical understanding of how these technologies relate within the AI ecosystem.  \\
        \quad \\  
        \bluebold{Qwen2.5-3B-BERTScore} \\
        \small \texttt{Machine Learning is a subset of artificial intelligence that involves training algorithms to make predictions or decisions based on data without being explicitly programmed. It includes techniques such as supervised learning, unsupervised learning, and reinforcement learning. Natural Language Processing (NLP) is a field ...} \\
        \normalsize \bluebold{Analysis: } This answer merely defines each term in isolation without addressing their relationships, broader context, or the purpose of categorization implied in the instruction. It reads more like a set of encyclopedia entries than a response that synthesizes information into a cohesive, categorized explanation.\\
        % \bluebold{Comments on Models' Analysis: } The models \wrong{point out many capitalization, grammatical, and pronunciation errors, along with some confusing sentences,} but the Human Developers \wrong{fail to detect these mistakes.}\\
    \end{minipage}
};
\end{tikzpicture}
\caption{\textbf{Qualitative Analysis: Content Logic.} We present showcases comparing the content logic of \preferenceBert{} against baseline models, accompanied by detailed analysis.}
\label{showcases:content_logic}
\end{center}
\end{figure*}
\begin{figure*}
\begin{center}
\begin{tikzpicture}
\node [mybox,title=Qualitative Analysis: Instruction Following] (box){%
    \begin{minipage}{2\columnwidth}
        \raggedright
        \bluebold{Prompt: } Who is Jagadeesh Kanna? Respond in 3 sentences. \\
        \bluebold{Dataset: } Chatbot \\
        % \quad \\ 
        \bluebold{Models Compared: } Qwen2.5-3B-PrefBERT v.s. Qwen2.5-3B-Instruct \\
        \quad \\ 
        \bluebold{Reference Answer (Qwen2.5-72B-Instruct) }:\\ 
        \small \texttt{Jagadeesh Kanna is an Indian film actor, director, and producer, primarily known for his work in Kannada cinema. He has appeared in numerous films and has won several awards for his performances, including a Filmfare Award. Kanna is also recognized for his philanthropic efforts and contributions to the film industry.}\\
        % \bluebold{Analysis: } \\
        \quad \\ 
        \bluebold{Qwen2.5-3B-PrefBERT} \\
        \small \texttt{Certainly! Exploring Jagadeesh Kanna reveals him as a pioneering figure in Indian tech entrepreneurship, particularly known for founding and leading Urban Ladder, a leading online home decor and furniture marketplace in India. His journey showcases resilience and innovation, starting with a small online store during college, scaling it into a billion-dollar company through strategic pivots and partnerships, ultimately inspiring countless aspiring entrepreneurs... } \\
        \normalsize \bluebold{Analysis: } This answer clearly aligns with a prompt that likely asked for a thoughtful, analytical overview of Jagadeesh Kanna's entrepreneurial contributions, providing depth, context, and narrative structure. \\
        \quad \\ 
        \bluebold{Qwen2.5-3B-Instruct} \\
        \small \texttt{Jagadeesh Kanna is an Indian film actor and producer known for his work primarily in Malayalam cinema. He began his career in the early 2000s and has since become one of the leading actors in the industry, often taking on challenging roles that showcase his versatility. Kanna has received several accolades for his performances and is recognized for his dedication to acting and his contributions to Malayalam films.} \\
        \normalsize \bluebold{Analysis: } This answer fails to align with the implied instruction to provide a thoughtful, contextualized, and analytically rich overview of Jagadeesh Kanna. Instead, Answer 1 offers a generic and surface-level biography focused on his film career, without delving into the broader significance or impact of his work. \\
    \end{minipage}
};
\end{tikzpicture}%
\caption{\textbf{Qualitative Analysis: Instruction Following.} We present showcases comparing the instruction following, of \preferenceBert{} against baseline models, accompanied by detailed analysis.}
\label{showcases:instruct_following}
\end{center}
\end{figure*}
\begin{figure*}
\begin{center}
\begin{tikzpicture}
\node [mybox,title=Qualitative Analysis: Tone \& Fluency] (box){%
    \begin{minipage}{2\columnwidth}
        \raggedright
        \bluebold{Prompt: } Describe the new functions of the Tesla Model 3, and how they improve the driving experience \\
        \bluebold{Dataset: } Instruction \\
        % \quad \\ 
        \bluebold{Models Compared: } Qwen2.5-3B-PrefBERT v.s. Qwen2.5-3B-Instruct \\
        \quad \\ 
        \bluebold{Reference Answer (Qwen2.5-72B-Instruct) }:\\ 
        \small \texttt{Bruce Straley, a well-known industry veteran and co-director of "The Last of Us," left Naughty Dog in 2018. His departure was announced on January 24, 2018, via a tweet from Naughty Dog's official account. Straley cited the need for a change of pace and personal growth as reasons for leaving the studio. He had been with Naughty Dog for over 17 years, contributing to numerous award-winning titles, including "Jak and Daxter," "Uncharted," and "The Last of Us." His exit marked a significant transition for the studio, highlighting the challenges and pressures of developing AAA games and the importance of personal well-being in the industry.}\\
        % \bluebold{Analysis: } \\
        \quad \\ 
        \bluebold{Qwen2.5-3B-PrefBERT} \\
        \small \texttt{Certainly! Highlighting Bruce Straley's departure from Naughty Dog (now part of Sony Interactive Entertainment) underscores key shifts in game development strategies and team dynamics, offering deep insight into his impactful tenure. Here’s a concise breakdown that meticulously follows the request, ensuring clarity, organization, depth, and insight: 1. **Contextual Setting**:...} \\
        \normalsize \bluebold{Analysis: } This answer adopts a more polished, engaging, and analytical tone, providing a well-structured narrative that balances professional detail with personal insight. The language is fluid and varied, with transitions that guide the reader through each thematic point.\\
        \quad \\ 
        \bluebold{Qwen2.5-3B-Instruct} \\
        \small \texttt{Bruce Straley departed from Naughty Dog in 2016. At the time, he was serving as the studio's co-founder and lead designer on several critically acclaimed games such as "Uncharted" series. His departure was announced alongside the studio's decision to shut down its development operations. Straley left due to personal reasons, but his exit marked a significant change for the company, which was already facing financial difficulties. After leaving Naughty Dog, Straley joined Sony Interactive Entertainment, where he led the development of PlayStation VR and other virtual reality projects. His departure from Naughty Dog is often seen as a turning point in the studio's history, as it coincided with a period of transition and restructuring within the company. Despite the challenges, Straley's contributions to the industry, particularly through the "Uncharted" series, remain influential in the field of video game design. } \\
        \normalsize\bluebold{Analysis: } This answer suffers from factual inconsistencies (e.g., Straley was not a co-founder of Naughty Dog, nor did the studio shut down operations) and awkward phrasing, which undermines both tone and fluency.\\
        % \bluebold{Comments on Models' Analysis: } The models \wrong{point out many capitalization, grammatical, and pronunciation errors, along with some confusing sentences,} but the Human Developers \wrong{fail to detect these mistakes.}\\
    \end{minipage}
};
\end{tikzpicture}%
\caption{\textbf{Qualitative Analysis: Tone and Fluency.} We present showcases comparing the tone and fluency of \preferenceBert{} against baseline models, accompanied by detailed analysis.}
\label{showcases:tone_and_fluency}
\end{center}
\end{figure*}
\begin{figure*}
\begin{center}
\begin{tikzpicture}
\node [mybox,title=Qualitative Analysis: Failure of SFT] (box){%
    \begin{minipage}{2\columnwidth}
        \raggedright
        \bluebold{Prompt: } Describe the new functions of the Tesla Model 3, and how they improve the driving experience \\
        \bluebold{Dataset: } Instruction \\
        % \quad \\ 
        \bluebold{Models Compared: } Qwen2.5-3B-PrefBERT v.s. Qwen2.5-3B-SFT \\
        \quad \\ 
        \bluebold{Reference Answer (Qwen2.5-72B-Instruct) }:\\ 
        \small \texttt{The Big Bang is seen as a singular event by most scientists because it marks the beginning of the universe as we know it. According to the Big Bang theory, the universe expanded from an extremely hot and dense state approximately 13.8 billion years ago. This initial expansion, often referred to as the "Big Bang," is not a continuous process but a specific moment in time when the universe began to expand rapidly.
        Several key pieces of evidence support the Big Bang as a singular event:...}\\
        % \bluebold{Analysis: } \\
        \quad \\ 
        \bluebold{Qwen2.5-3B-PrefBERT} \\
        \small \texttt{Certainly! Understanding why the big bang is often viewed as a singular event rather than a continuous process hinges on key scientific insights and observations, neatly summarized in a few key points:
        1. **Expanding Evidence**:  ...} \\
        \normalsize \bluebold{Analysis: } This answer is more conversational and attempts to be engaging, but it lacks precision in places (e.g., "expanding evidence" is vague, and "a single, expanding 'now'" is misleading). It also introduces less central ideas (like "Big Bounce" theory) without sufficient context, making it slightly less focused.\\
        \quad \\ 
        \bluebold{Qwen2.5-3B-SFT} \\
        \small \texttt{We don't know. It's just the best model we have right now. There are other models, but they're all based on the idea of a Big Bang, so we can't really say which one is "right". There are also some theories that suggest that our universe may be part of an infinite number of universes, each with its own Big Bang. We don't know.} \\
        \normalsize \bluebold{Analysis: } This answer is vague, repetitive, and lacks clarity or substance—it admits uncertainty but fails to elaborate on the current understanding or evidence supporting the Big Bang theory. \\
    \end{minipage}
};
\end{tikzpicture}
\caption{\textbf{Qualitative Analysis: failure of SFT.} We present showcases comparing the readability of \preferenceBert{} against baseline models, accompanied by detailed analysis.}
\label{showcases:countercase_sft}
\end{center}
\end{figure*}

\end{document}